\pdfoutput=1

\documentclass[11pt]{article}

\usepackage{acl}

\usepackage{booktabs}
\usepackage{array}
\usepackage{lipsum}
\usepackage{amsmath}        
\usepackage{amssymb}        
\usepackage{mathtools}
\usepackage{subcaption}
\usepackage{times}
\usepackage{latexsym}
\usepackage{CJKutf8}
\usepackage[T1]{fontenc}
\usepackage[utf8]{inputenc}

\usepackage{hyperref}
\usepackage{url}
\usepackage{booktabs}
\usepackage{color}
\usepackage{xcolor}
\definecolor{lightgreen}{RGB}{126,217,87}
\PassOptionsToPackage{prologue,dvipsnames}{xcolor}
\usepackage{colortbl}
\usepackage[dvipsnames]{xcolor}
\usepackage{lineno}

\usepackage{commenting}
\usepackage{tcolorbox}
\usepackage{multirow}
\usepackage{amsmath}
\usepackage{rotating} 
\usepackage{array}
\usepackage{tabularx}

\definecolor{darkblue}{rgb}{0, 0, 0.5}
\hypersetup{colorlinks=true, citecolor=darkblue, linkcolor=darkblue, urlcolor=darkblue}

\usepackage{algorithm}
\usepackage{algpseudocode}
\usepackage{graphicx}
\usepackage{makecell}
\usepackage{fontawesome}
\usepackage{caption}
\usepackage{subcaption}

\title{Mitigating Behavioral Hallucination in Multimodal Large Language Models for Sequential Images}

\author{
Liangliang You\textsuperscript{*,1},
Junchi Yao\textsuperscript{*,1,3},
Shu Yang\textsuperscript{1,2},
Guimin Hu\textsuperscript{4},
Lijie Hu\textsuperscript{†,1,2},
Di Wang\textsuperscript{†,1,2}\\
$^1$Provable Responsible AI and Data Analytics (PRADA) Lab\\
$^2$King Abdullah University of Science and Technology \\
$^3$University of Electronic Science and Technology of China
$^4$University of Copenhagen
}

\begin{document}
\maketitle

\begin{abstract}
While multimodal large language models excel at various tasks, they still suffer from hallucinations, which limit their reliability and scalability for broader domain applications. To address this issue, recent research mainly focuses on objective hallucination. However, for sequential images, besides objective hallucination, there is also behavioral hallucination, which is less studied. This work aims to fill in the gap. We first reveal that behavioral hallucinations mainly arise from two key factors: prior-driven bias and the snowball effect. Based on these observations, we introduce SHE (Sequence Hallucination Eradication), a lightweight, two-stage framework that (1) detects hallucinations via visual-textual alignment check using our proposed adaptive temporal window and (2) mitigates them via orthogonal projection onto the joint embedding space. We also propose a new metric (BEACH) to quantify behavioral hallucination severity. Empirical results on standard benchmarks demonstrate that SHE reduces behavioral hallucination by over 10\% on BEACH while maintaining descriptive accuracy.
\end{abstract}

\def\thefootnote{*}\footnotetext{Equal Contribution.}
\def\thefootnote{†}\footnotetext{Corresponding Author.}

\section{Introduction}
\label{sec:intro}
Large language models (LLMs) have demonstrated powerful vision--language understanding and generation capabilities across various domains~\cite{su2023detectllm,yang2024monal,xu2023llm,zhang2025mechanistic,yang2025fraud,cheng2025codemenv,hu2024understanding}. Recently, 
multimodal large language models (MLLMs), such as GPT-4V~\citep{yang2023dawn} and Qwen2-vl~\citep{wang2024qwen2vl}, have been developed to extend to the visual modality. In particular, they achieve outstanding performance on tasks such as visual question answering~\citep{ye2023mplugowl2}, visual reasoning~\citep{liu2024rar}, video summarization~\citep{hua2025v2xum}, and image captioning~\citep{chen2024compcap}. Despite their remarkable successes, these models still suffer from inherent limitations that lead to hallucinations, i.e., generating text that is semantically coherent but inconsistent with the visual content. Such errors raise critical reliability concerns, especially in sensitive domains like education, law, and medicine, where they can distort learning outcomes~\citep{wang2024mementos, li2023evaluating, yin2023survey,cheng2024multi,ali2024mqa,zhang2024locate,cheng2025compke,yang2024makes}.

\begin{figure*}[t]
    \centering
    \includegraphics[width=0.85\textwidth]{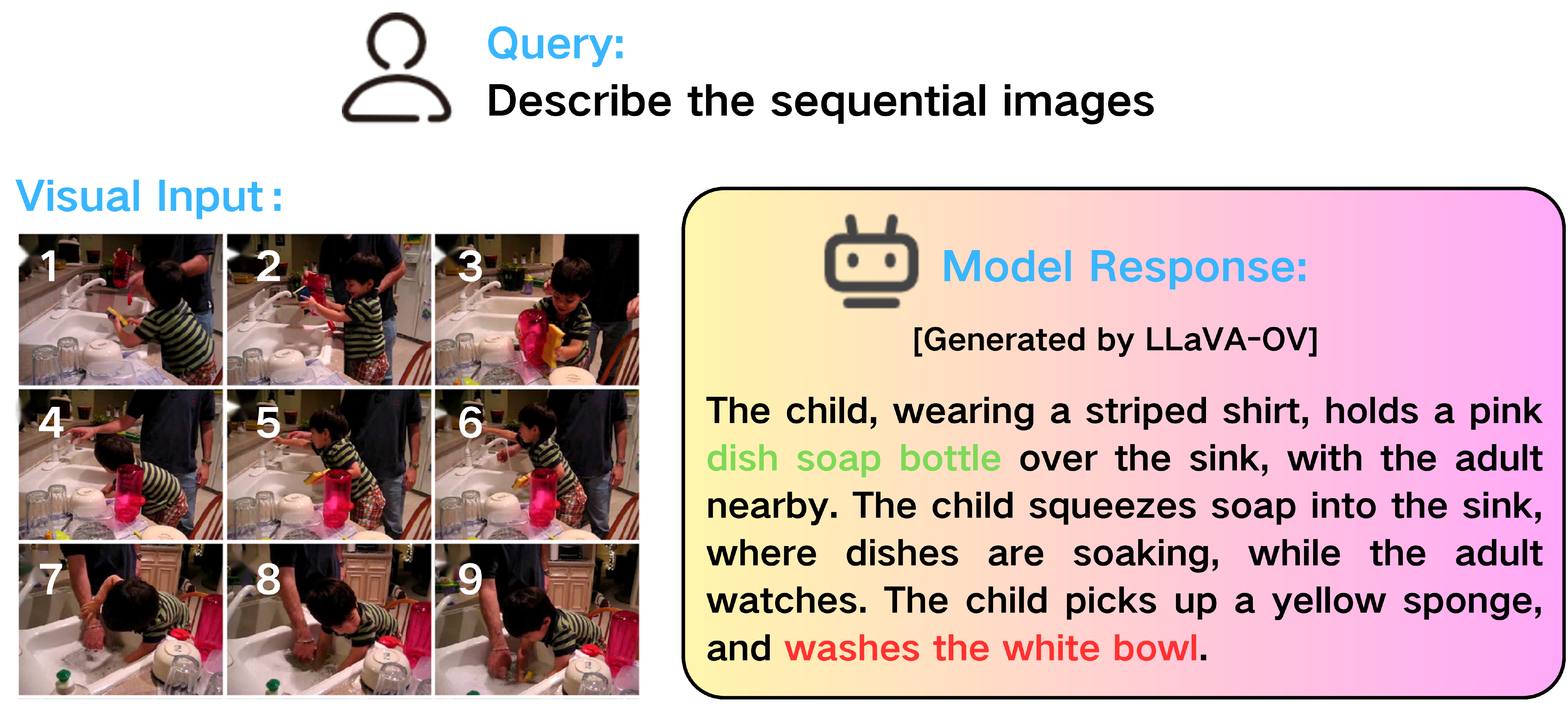}
    \caption{{\color{red} \textbf{Behavior hallucinations}} indicate that the model generated behaviors not present in the image. There is no washing of a white bowl in the image; the bowl is simply placed on the table. {\color{lightgreen} \textbf{Object hallucinations}} indicate that the output contains objects that are wrong or do not exist.}
    \label{fig:intro}
\vspace{-12pt}
\end{figure*}

Recent research has made significant progress in addressing object hallucinations in MLLMs for single images through methods like knowledge grounding and representation learning~\cite{yu2023hallucidoctor,jain2023vcoder,benkish2023mocha,zhao2023beyond}. However, when the input is a sequence of images, besides object hallucinations, there are behavioral hallucinations, which are more challenging and less studied. In detail, object hallucination occurs when the model describes objects in images that are not actually there, making the output not match the real image. Behavioral hallucination is when a model makes up behaviors for objects in images that they should not be doing based on what's actually shown (see Fig.~\ref{fig:intro} for an example). Unlike object hallucinations, these dynamic inconsistencies involve implausible actions or interactions between objects across temporal frames. 
To mitigate behavioral hallucinations, several approaches have been developed recently, such as Volcano \cite{lee2023volcano} and Self-PEP \cite{wang2024videohallucer}. However, these methods face limitations in either computational efficiency or dependency on external models; these would necessitate training external models or fine-tuning the original models, both of which greatly increase computational costs. 

To address these gaps, we propose SHE (\underline{\textbf{S}}equence \underline{\textbf{H}}allucination \underline{\textbf{E}}radication), a novel framework that detects and mitigates behavior hallucinations through latent representations. To motivate our method, we first reveal and analyze the prior-driven and snowball effects on generating behavioral hallucinations. Generally speaking, prior-driven mechanisms in MLLMs cause hallucinations by making the model rely too much on its existing biases from training, which skews its interpretation of inputs and leads to outputs that do not match reality. The snowball effect occurs when a single incorrect action prediction becomes the anchor that the model folds back into its context, triggering a cascading accumulation of additional related hallucinated behaviors.

Based on our analysis, SHE involves hallucination detection and mitigation stages. We detect hallucinations by checking how well patches of images match the text they generate. To further leverage the snowball effect, we propose an adaptive temporal windowing for the hallucination detection. Then, SHE involves using contextual embeddings to spot inconsistencies and then applying orthogonal projection in the vision embedding space to eliminate the hallucinated elements.  By maintaining the consistency between visual and textual information without needing additional training, it enhances the reliability of MLLMs.  

Compared to other methods, SHE shows superior performance across multiple evaluation metrics. Specifically,  we propose an evaluation metric, namely BEACH, to better assess behavioral hallucinations, and show SHE achieves a 10\% reduction in BEACH. These results confirm the effectiveness and practicality of our approach, offering a novel solution to the behavior hallucination problem in MLLMs for sequential images. 
Our contributions are threefold:

\begin{enumerate}
    \item We provide comprehensive studies on the causes of behavior hallucinations and identify two key factors: the prior-driven effect and the snowball effect. We also analyze the correlation and difference between the influence of these two causes.
    \item Based on our analysis, we propose SHE, a two-stage framework for addressing MLLM behavior hallucinations. SHE first detects behavior hallucinations using embeddings of our proposed adaptive temporal window, and then corrects them via projecting representations without extra training.
    \item Our work addresses a research gap on behavioral hallucinations, with experiments showing that our method's relief effectiveness surpasses baseline approaches. The results also demonstrate that SHE can effectively manage behavioral hallucinations while maintaining descriptive accuracy.
 
\end{enumerate}

\section{Related Work}
Research on hallucination mitigation in multimodal models spans four axes, i.e., data-level methods (Llava-1.5’s negative instruction tuning~\cite{liu2023mitigating}, caption rewriting~\cite{wang2023recaption}), model-centric enhancements (InternVL’s high-resolution grounding~\cite{chen2023intervl}, HACL’s contrastive alignment~\cite{zhao2023reca}), dataset calibration via HalluciDoctor’s~\cite{yu2023hallucidoctor} cleaned instructions and counterfactual visual instructions to expand dataset diversity, and caption refinement with ReCaption’s rewritten in existing datasets to produce high-quality image–caption pairs~\cite{xing2024efuf}.

Training innovations include RLHF-based methods like ViGoR~\cite{yan2024vigor} and frameworks such as MOCHa~\cite{benkish2023mocha}, which use dense rewards to align captions with human preferences and boost accuracy. Other RL approaches (HA-DPO~\cite{zhao2023beyond}, RLHF techniques~\cite{sun2023aligning}) prioritize factual responses to curb hallucinations. At inference, decoding strategies VCD~\cite{leng2024mitigating}, HALC~\cite{chen2024halc}, IBD~\cite{zhu2024ibd}, and guided methods like GCD~\cite{deng2024seeing} dynamically anchor generation to visual input, ensuring consistency. However, these techniques primarily target static object hallucinations through frame-wise processing, neglecting temporal dynamics in sequential inputs due to two gaps: (1) lack of inter-frame dependency modeling for error propagation control, and (2) reliance on multi-stage pipelines incurring prohibitive latency (e.g., Woodpecker~\cite{yin2023woodpecker} five-step correction). Thus, this prompts video-focused solutions to attempt temporal modeling, such as Volcano~\cite{lee2023volcano}, VISTA-LLAMA~\cite{ma2023vistallama}, MERLIM~\cite{villa2023behind}.

Our research focuses on temporal hallucinations in image sequences, which are markedly distinct from existing methods. Most traditional approaches, such as LRV-Instruction, InternVL, and VCD, concentrate on static object hallucinations by processing frames independently and thus overlook inter-frame dependencies, temporal dynamics, and effective use of visual context. Although recent video-oriented methods like Volcano and VISTA-LLAMA incorporate temporal information, they rely on reactive corrections and do not address the root causes of hallucinations.

\section{Why MLLMs Generate Hallucinations?}
\label{sec:causal_analysis}
Before detailing our proposed method, we first examine why MLLMs generate behavior hallucinations. Specifically, we identify two causes: the prior-driven effect and the snowball effect. Generally, the prior-driven effect induces hallucinations by causing the model to rely excessively on its pre-existing biases, which skews its interpretation of inputs and leads to outputs that do not match reality. The snowball effect in hallucinations occurs when a single incorrect action prediction becomes the anchor that the model folds back into its context, triggering a cascading accumulation of additional related hallucinated behaviors. We discuss these effects in the following subsections. 

\subsection{Prior-Driven Effects}
\label{sec:experimental_setup}

To study the prior-driven effect on hallucinations, we examine how hallucinated behaviors are contextually linked to non-hallucinatory behaviors or hallucinatory objects, thereby revealing the model’s prior bias in generating related hallucinations. We quantify this alignment using the co-occurrence score, which measures the strength of association between two elements by capturing how frequently they appear together within the same context~\cite{kang2023impact}. Intuitively, a high co-occurrence score between the generated response and the input image indicates that the model filled in missing or ambiguous visual details based on learned textual associations instead of relying on truly observed objects or behaviors. 
Thus, by measuring how often hallucinatory behaviors co-occur with specific objects or other behaviors, we directly capture the magnitude of the model’s prior-driven bias. 

\begin{figure*}[t]
\vspace{-7pt}
    \centering
    \begin{subfigure}{0.45\textwidth}
        \centering
        \includegraphics[width=0.99\textwidth]{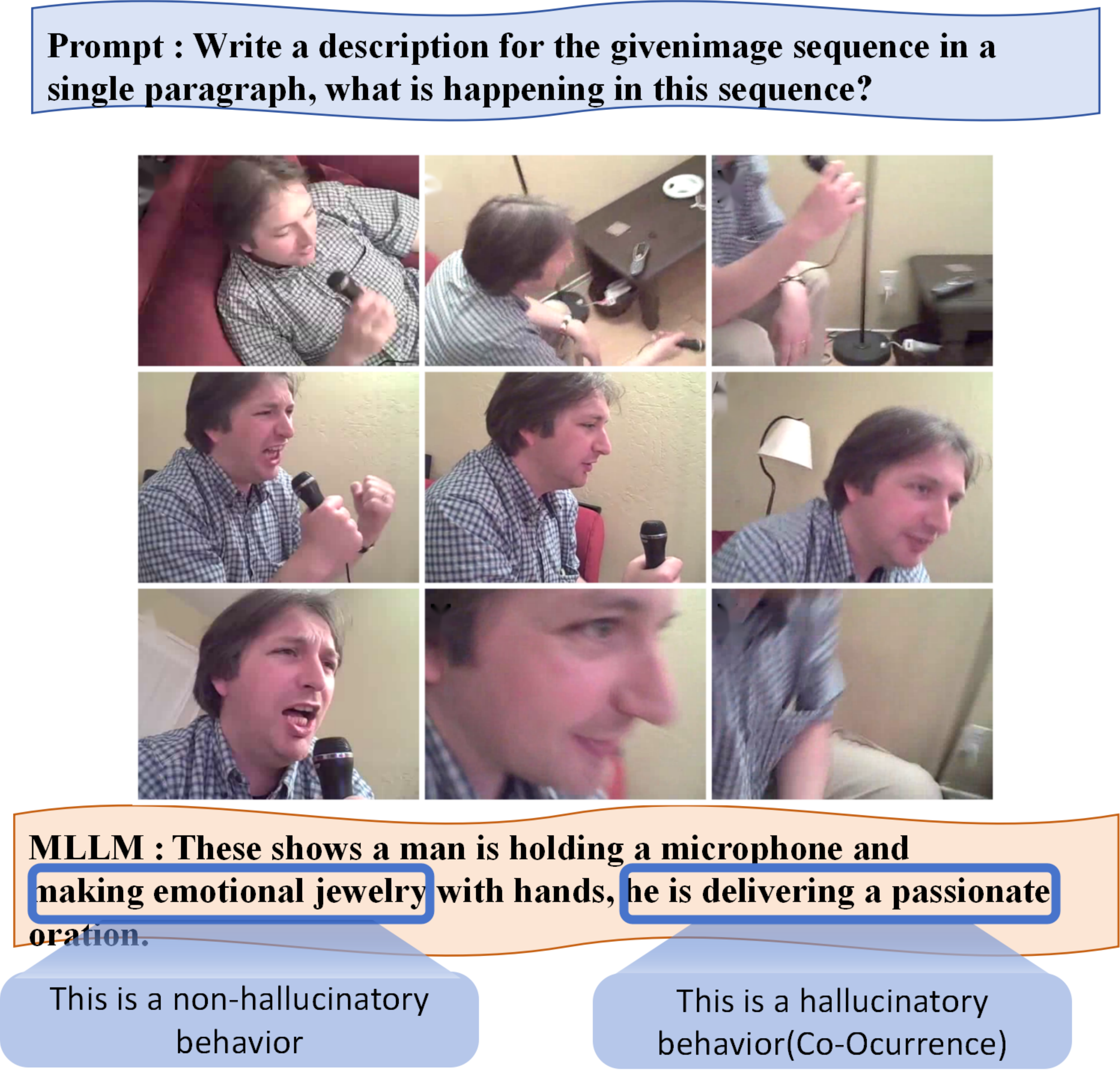}
        \captionsetup{width=0.9\textwidth}
        \caption{An example of behavior-to-hallucination behavior co-occurrence. }
        \label{3.1 exam(a)}
    \end{subfigure}
\begin{subfigure}{0.45\textwidth}
        \centering
        \includegraphics[width=0.99\textwidth]{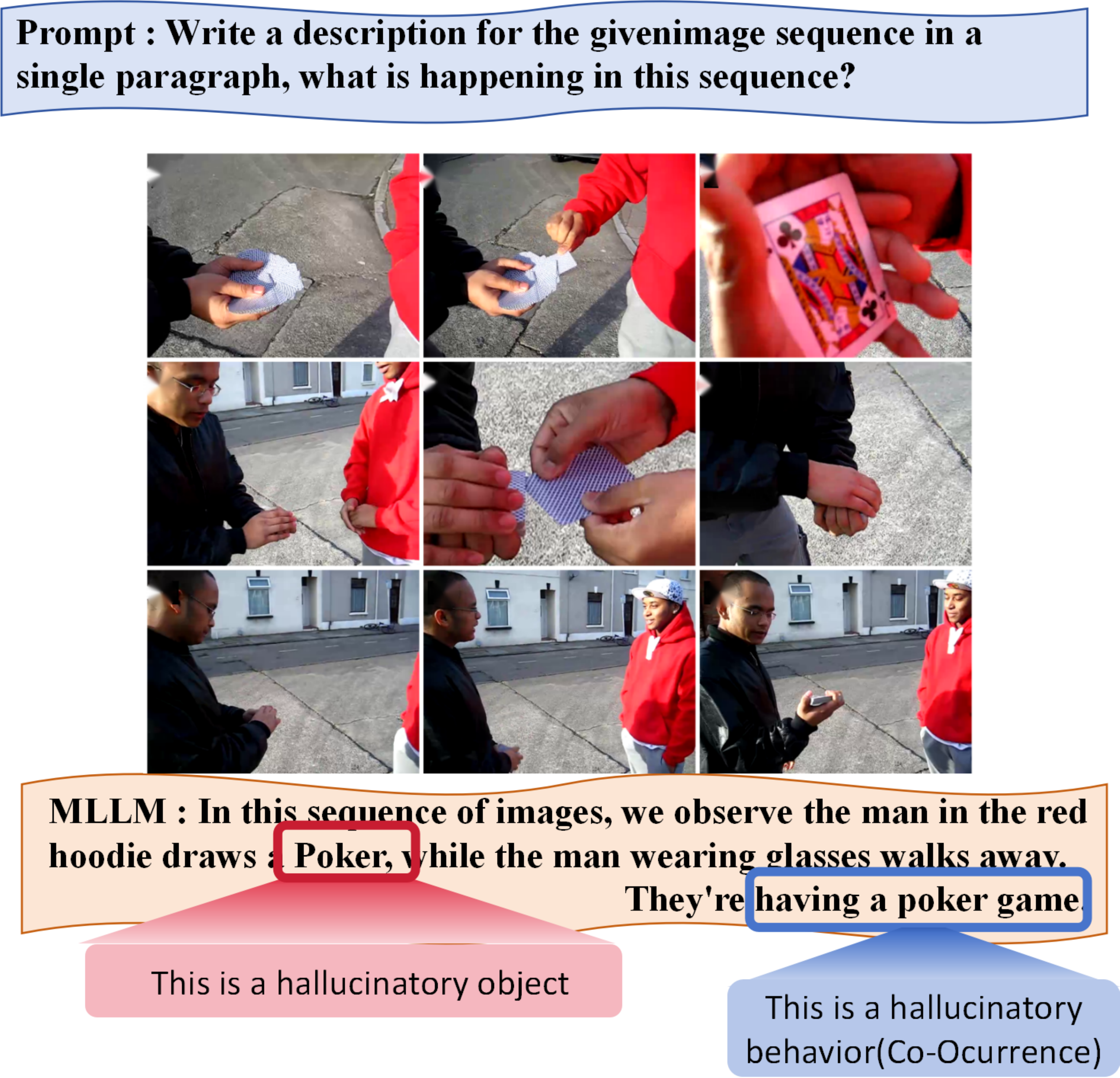}
        \captionsetup{width=0.9\textwidth}
        \caption{An example of hallucination behavior-to-object co-occurrence.}
        \label{3.1 exam 2(b)}
    \end{subfigure}
    \caption{\textbf{(a)} The model correctly recognizes “making emotional jewelry” (highlighted in red), but then hallucinates by generating “holding a microphone.” \textbf{(b)} The model correctly identifies “Poker” (highlighted in blue), yet hallucinates by producing “have poker gaming.” }
    \label{Fig 1}
\vspace{-12pt}
\end{figure*}

However, currently there are no metrics that are tailored for evaluating co-occurrence either between behaviors themselves or between behaviors and objects. To address this gap, we propose two novel metrics: one quantifies behavior-to-hallucinated behavior co-occurrence we call it \textbf{C}o-\textbf{O}ccurence \textbf{S}core (\textbf{B}ehavior-to-\textbf{H}allucination Behavior) $\mathrm{CoS}(\mathrm{BH})$, and another measures hallucinated behavior-to-object co-occurrence we call it \textbf{C}o-\textbf{O}ccurence \textbf{S}core (Hallucination {B}ehavior-to-\textbf{O}bject) $\mathrm{CoS}(\mathrm{BO})$. 
For the hallucinated and descriptive caption $c$ generated by an MLLM for a set of sequence images, we define its $\mathrm{CoS}(\mathrm{BH})$ and $\mathrm{CoS}(\mathrm{BO})$ as the following.  A larger value generally means more frequent hallucinations.
\begin{align*}
\mathrm{CoS_c}(\mathrm{BH}) &= \sum_{i=1}^{n_h} \sum_{j=1}^{ n_r} \frac{|C(b_{c,i}) \cap C(b_{c,j})|}{|C(b_{c,i})| + |C(b_{c,j})|}, 
\end{align*}
where \( b_{c,i} \) is the $i$-th hallucinatory behavior and $b_{c,j}$ is the $j$-th real behavior in caption $c$, \( C(b_{c,i}) \) is the set of all captions mentioning behavior \( b_{c,i} \), \( n_h \) is the number of hallucinatory behaviors, and  \( n_r \) is the number of real behaviors. Generally speaking, $\mathrm{CoS_c}(\mathrm{BH})$ measures how often the hallucinated behaviors co-occur with non-hallucinated ones.  See Fig.~\ref{3.1 exam(a)} for an example. 
\begin{align*}
\mathrm{CoS_c}(\mathrm{BO}) &= \sum_{i=1}^{n_h} \sum_{j=1}^{m} \frac{|C(b_{c,i}) \cap C(o_{c,j})|}{|C(b_{c,i})| + |C(o_{c,j})|},
\end{align*}
where \( o_{c,j} \) is the $j$-th hallucinatory object in caption $c$, \( O(o_{c,j}) \) is the set of all captions mentioning object \( o_{c,j} \), and \( m \) is the number of hallucinatory objects in the object set. Unlike $\mathrm{CoS_c}(\mathrm{BH})$, $\mathrm{CoS_c}(\mathrm{BO})$ checks how well the model links behaviors to hallucinatory objects in a logical way that matches the context. See Figure~\ref{3.1 exam 2(b)} for an example.

For comparison, we also define corresponding scores for captions of the same image sequence that contain no hallucinatory behaviors. We sample \( n_h \)  non-hallucinatory behaviors to match the hallucination group and let \( n_r \) and \( m \) be the same as before.

In our experiment, we consider Llava-OneVsion~\cite{li2024llavaonevision} as the MLLM and  the Mementos~\citep{wang2024mementos} dataset.
We start by using  Llava-OneVision to generate captions for each image sequence in the dataset. 

\begin{figure*}[ht]
    \centering
    \begin{subfigure}{0.45\textwidth}
        \includegraphics[width=0.8\textwidth]{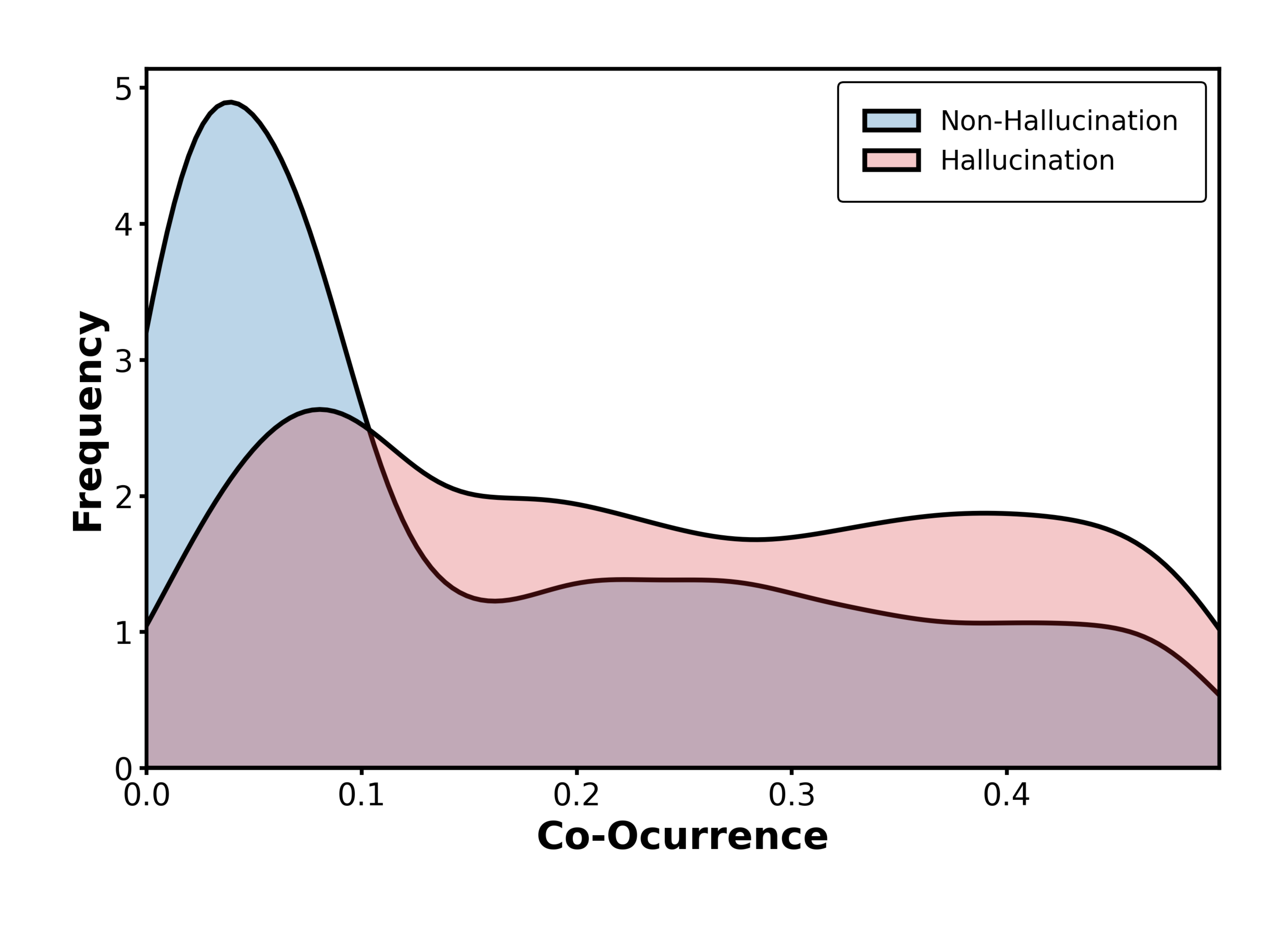}
        \captionsetup{width=0.99\textwidth, font=footnotesize }
        \vspace{-12pt}
        \caption{$\mathrm{CoS_c}(\mathrm{BH})$}
        \label{Fig 2(a)}
    \end{subfigure}
    \hfill
    \begin{subfigure}{0.45\textwidth}
        \includegraphics[width=0.8\textwidth]{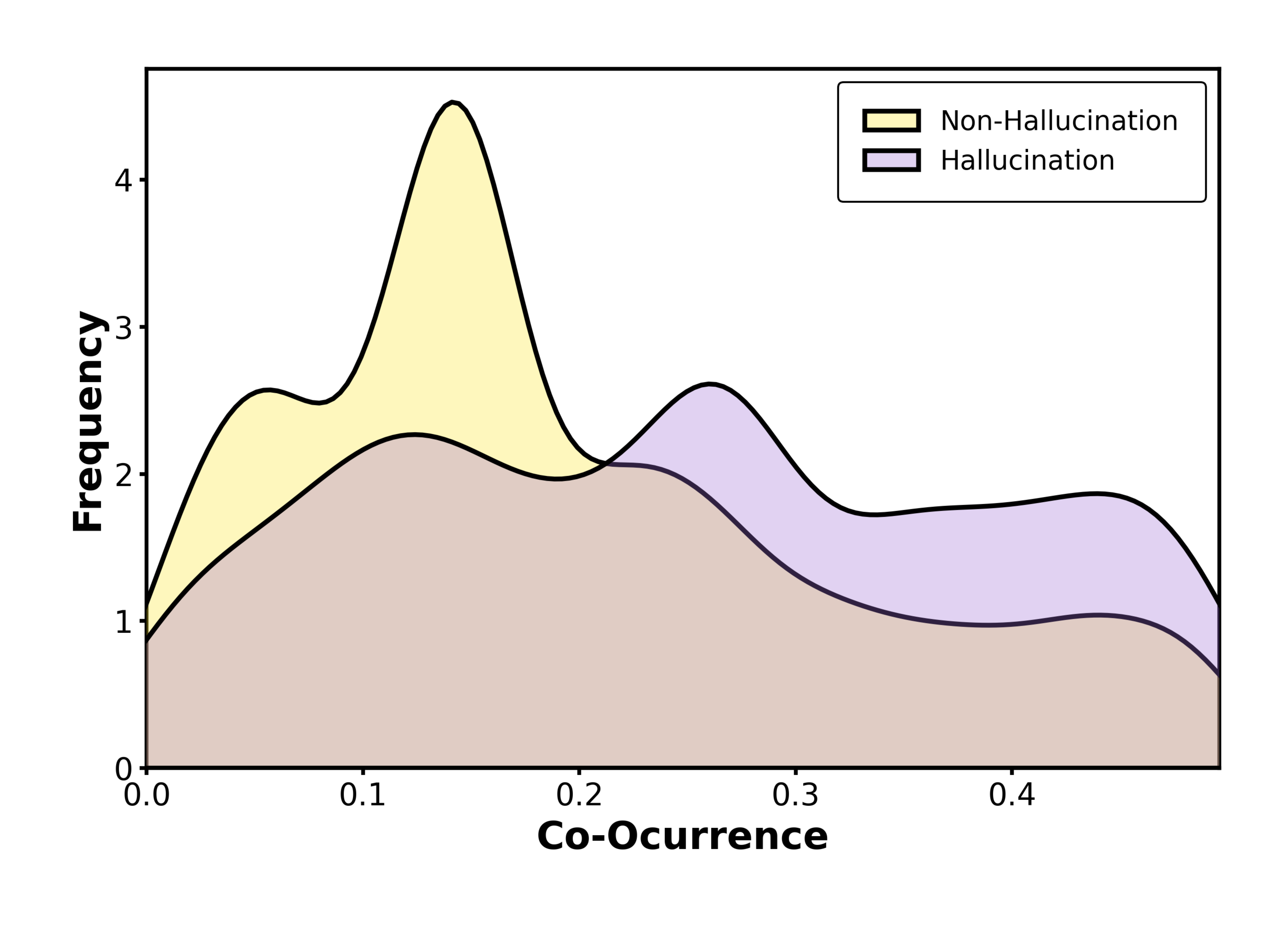}
        \captionsetup{width=0.99\textwidth, font=footnotesize }
         \vspace{-12pt}
        \caption{$\mathrm{CoS_c}(\mathrm{BH})$
    }
        \label{Fig 2(b)}
    \end{subfigure}
         \vspace{-7pt}
        \caption{ Distribution of $\mathrm{CoS_c}(\mathrm{BH})$ and $\mathrm{CoS_c}(\mathrm{BH})$ for hallucinatory  and non-hallucinatory captions.  
}
	\label{Fig 2}
\vspace{-12pt}
\end{figure*}
\noindent {\bf Results Analysis.}
Prior-driven effects indicate that MLLMs are sometimes driven by or influenced by their prior knowledge rather than carefully checking what is actually in the images. 
To quantify such prior bias, we use $\mathrm{CoS}(\mathrm{BH})$ and $\mathrm{CoS}(\mathrm{BO})$ to measure the extent to which the model’s hallucinations reflect learned associations. 

In Figure~\ref{Fig 2(a)}, captions containing behavioral hallucinations exhibit a pronounced bias: non-hallucinated captions mostly score between  $0.05$ and $0.15$ while hallucinated ones are beyond $0.4$. Moreover, Figure~\ref{Fig 2(b)}, 
shows that the scores of non-hallucinated captions concentrate between $0.1$ and $0.2$ while hallucinated ones  are more divergent. These patterns arise from the model’s reliance on prior knowledge acquired during training. During training, the model learns associations between behaviors, objects, and textual descriptions, which form its prior knowledge. When encountering unseen inputs without sufficient visual evidence, the model often defaults to these learned associations, leading to hallucinated behaviors. For example, if the model frequently associates a behavior with a specific object during training, it is more likely to hallucinate that behavior when the object is present. Moreover, the model’s hallucinations also reflect learned object–behavior relationships from the training data.

Next, we demonstrate prior bias through attention distribution, which refers to the allocation of the model’s focus between different input modalities, such as text and visual tokens.  Attention analysis in Appendix Fig.~\ref{Fig 3} shows that 87\% of the model’s attention is allocated to text tokens, despite visual tokens comprising nearly 90\% of the total tokens. This imbalance underscores the model’s reliance on prior knowledge and amplifies the prior-driven effects, thereby contributing to hallucinations. 
\subsection{Snowball Effects }
\label{sec:Snowball}
\begin{figure*}[htbp]
\vspace{-7pt}
    \centering
    \begin{subfigure}{0.325\textwidth}
    	\centering
        \includegraphics[width=0.9\textwidth]{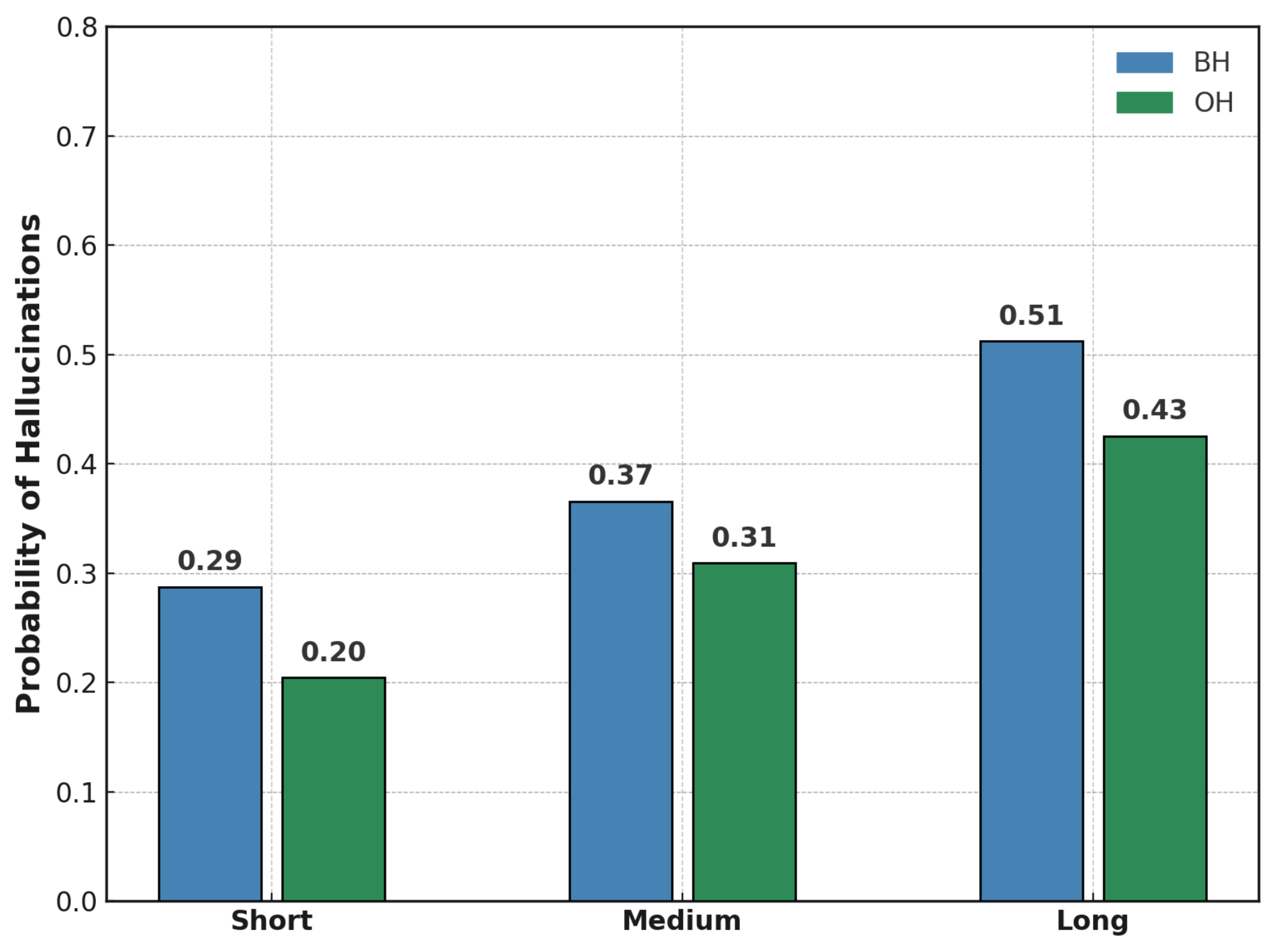}
        \caption{Video Length}
        \label{Fig 4(a)}
    \end{subfigure}
    \hfill
    \begin{subfigure}{0.325\textwidth}
    	\centering
        \includegraphics[width=0.9\textwidth]{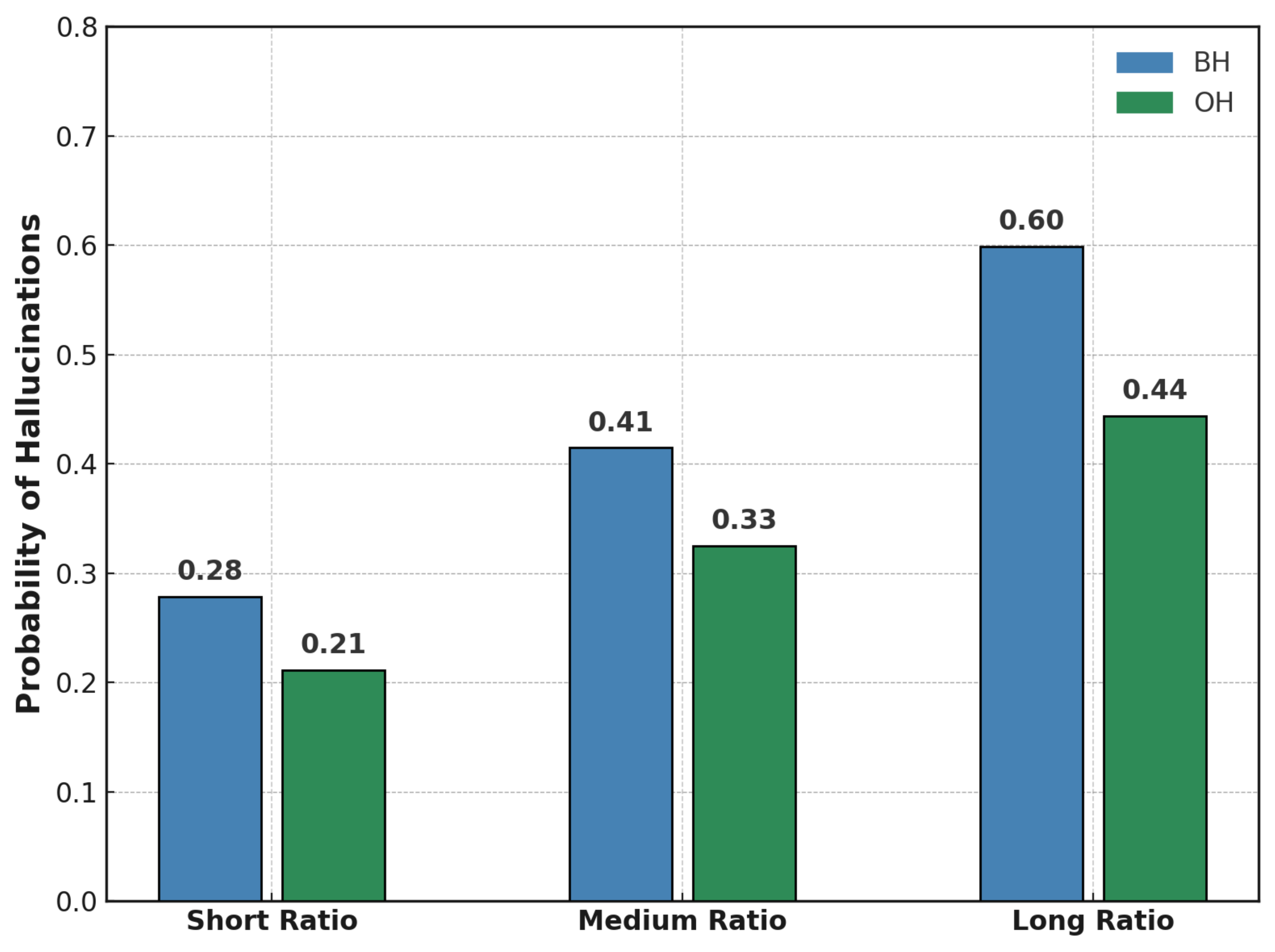}
        \caption{ Sequence Length}
        \label{Fig 4(b)}
    \end{subfigure}
    \begin{subfigure}{0.325\textwidth}
    	\centering
        \includegraphics[width=0.9\textwidth]{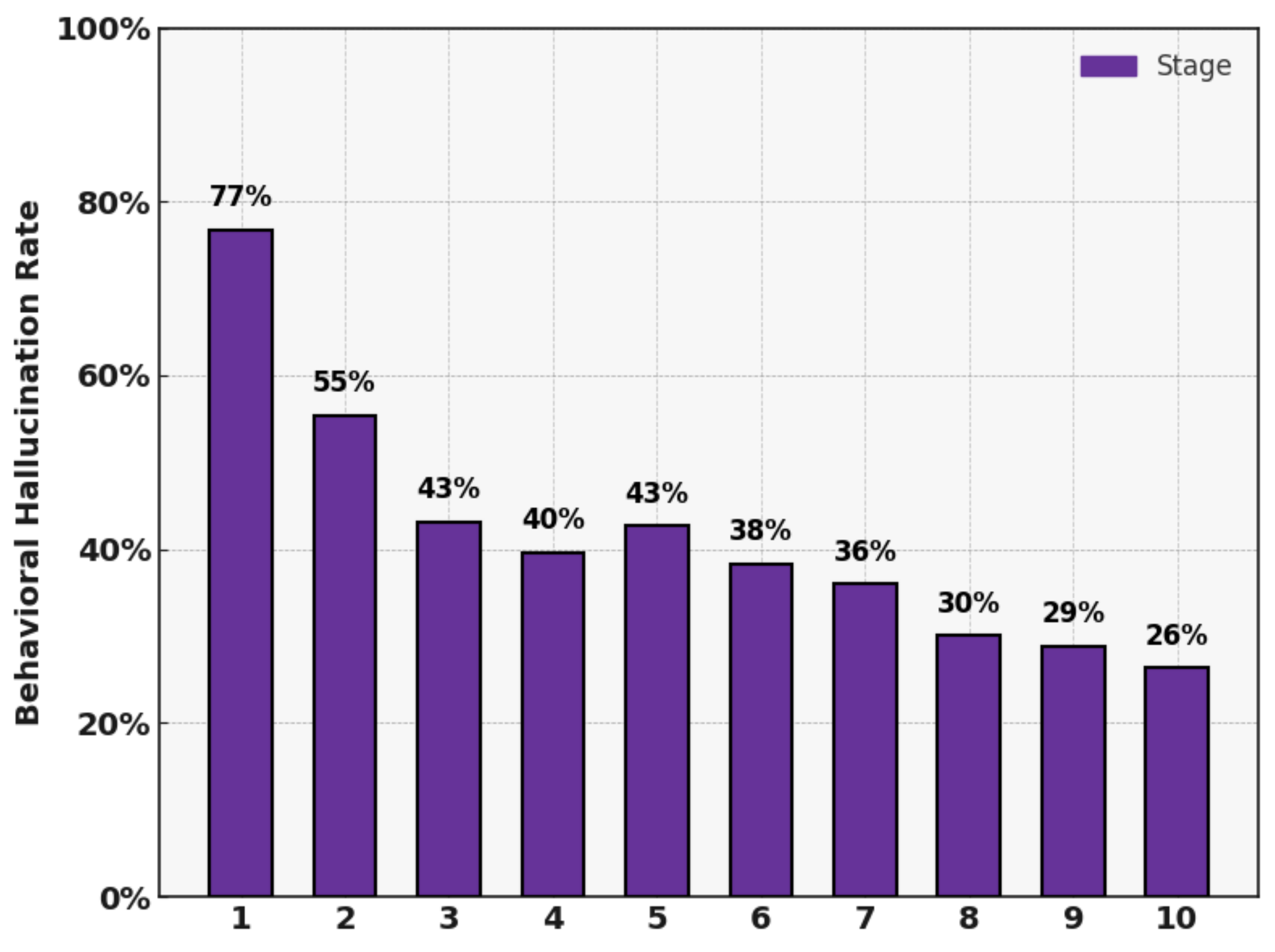}
        \caption{Perturbation Experiment}
        \label{Fig 4(c)}
    \end{subfigure}
    \caption{\textbf{(a)} Distribution of hallucination rates for behavioral hallucination (BH)  and objective hallucination(OH) with different video lengths. \textbf{(b)} Distribution of hallucination rates for 
 BH and OH  for different lengths of the same videos. \textbf{(c)} Behavioral hallucination rates after perturbing frames at different stages. }
	\label{Fig 4}
\vspace{-12pt}
\end{figure*}

To examine this effect, we use the Video-MME~\citep{fu2024video} dataset. After filtering out low-quality or incompletely annotated videos, we categorize the remaining samples by duration: short (<2 min), medium (2–10 min), and long (>10 min). For each category, we compute the behavior hallucination rate (BH) and object hallucination rate (OH) as the ratio of hallucinated responses to the total number of responses. Figure~\ref{Fig 4(a)} shows a clear increase in hallucination rate with increasing video length, indicating a positive correlation between sequence duration and hallucination prevalence.

Next, we assess the effect of sampling density. For each category, we extract frames at 1, 2, and 3 fps and compute the behavior hallucination rate (BH) as the ratio of hallucinated responses to the total number of responses. Figure~\ref{Fig 4(b)} shows shows that longer sequences and higher sampling rates lead to increased hallucination rates, reinforcing the observed correlation and motivating our subsequent controlled perturbation experiments.

To further explore the relationship between behavioral hallucinations and cumulative snowball effects, we perform stage-wise perturbations on each sequence. We divide each image sequence into ten equal segments \(s_i\) (\(i=1,\dots,10\)) and introduce either Gaussian blur (\(\sigma=1.5\))~\citep{zhang2024vl} or 20\% random occlusion~\citep{wu2024autohallusion} to each segment. We then measure the change in behavior hallucination rate. Figure~\ref{Fig 4(c)} shows that perturbing the first segment (the first 10\% of frames) yields the largest increase in hallucinations, confirming that early errors disproportionately propagate and amplify through the sequence.

To summarize, our results show that hallucinations increase rapidly as the input sequence lengthens, especially for behavioral hallucinations. This trend occurs because the model processes each frame sequentially: an early mistake leads to more mistakes later. For objective hallucinations, errors are static—for example, the model might label a ``chair'' as a ``table'' in every frame. In contrast, behavioral hallucinations evolve over time: misrecognizing one action can trigger a cascade of unlikely behavior predictions that intensify throughout the sequence. Stage-specific perturbation experiments in Figure~\ref{Fig 4(c)} further illustrate this effect, showing that disturbances in early frames have the greatest impact. Noise injected into the first 10\% of frames accounts for 76.7\% of all behavioral hallucination perturbations in the middle frames contribute 42\%, and those in the final frames have negligible impact. These findings underscore the critical influence of initial processing stages: errors in the first few frames heavily bias subsequent predictions, especially for complex behavioral tasks.

\section{SHE: Sequence Hallucination Eradication}
In this section, we detail our methodology.  SHE concurrently addresses both hallucination mechanisms: for prior-driven effects caused by knowledge–visual misalignment, we detect them by measuring similarity between visual features and textual captions; to handle snowball effects, we employ adaptive temporal windows. To mitigate these effects,  we eliminate spurious associations via orthogonal projection in the joint embedding space, enforcing visually grounded reasoning while preserving linguistic coherence. 

\subsection{Detection}
\noindent\textbf{Visual-Textual Alignment Check.} 
Motivated by the finding in Section~\ref{sec:experimental_setup}  that visual tokens receive little attention—MLLMs focus less on visual evidence—we quantify semantic congruence between language captions and visual inputs.
We measure the similarity between contextual embeddings and image patch embeddings to detect hallucinations.
For each generated caption, we extract contextual embeddings of behavior tokens from intermediate layers and compute the global behavior representation \( e_{Beh} \) by averaging:
\begin{align*}
e_{Beh} = \frac{1}{n} \sum_{i=1}^{n} e_{l_T}(t_i),
\end{align*}
where \(\{t_1, t_2, \ldots, t_n\}\) are the tokens for representing the behavior to be detected, and \( e_{l_T}(t_i) \) denotes the embedding of token \( t_i \) at layer \( l_T \) ( layer selection is discussed in the experimental part). For each frame in the image sequence, we divide the image into patches and extract embeddings \( e_{l_I}(p_j) \) for each patch \( p_j \).

We then compute the cosine similarity between the behavior embedding \( e_{Beh} \) and each patch's embedding \( e_{l_I}(p_j) \) in every layer:
\begin{equation*}
\mathrm{Score}(p_j)
= \max_{\,l_I \in \{1,2,\dots,L\}}
\cos\bigl(e_{\mathrm{Beh}},\,e_{l_I}(p_j)\bigr).
\end{equation*}
This score quantifies semantic consistency between image patches and the behavior in the spatio-temporal domain. For each frame $f$, we aggregate its patch scores to compute a frame-level match metric. We then aggregate these metrics across all frames—typically using the maximum patch score across all frames as the global confidence score:
\begin{equation*}
\text{Confidence} = \max_{f \in \text{frames}, \, p_j \in f} \text{Score}(p_j). 
\end{equation*}
Confidence scores reflect the degree to which a behavior is grounded in visual evidence. Behaviors with higher scores are regarded as non-hallucinated, while those with lower scores are considered likely to be hallucinated.

\noindent\textbf{Adaptive Temporal Windowing.} 
To mitigate the accumulation of hallucinations over time, we propose adaptive temporal windowing to address the snowball effect. Specifically, for each image patch \( p_j^t \) requiring correction, we aggregate features from adjacent frames, thereby enhancing the robustness of our method:
\[
E_{\text{agg}}(p_j^{t}) = \{ e_{l_I}(p_j^{t-{\tau}/2}), ..,e_{l_I}(p_j^t), ..,e_{l_I}(p_j^{t+{\tau}/2}) \}, 
\]
where $\tau$, the temporal window radius determined by behavior complexity, is adaptively scaled by the entropy of $e_{\text{Beh}}$. This adaptive scaling ensures that embeddings with higher semantic uncertainty receive wider windows to capture additional spatio-temporal context and mitigate cascading hallucinations. Specifically, we define:
  \[
  \tau = \lceil \ \gamma \cdot \text{Entropy}(e_{\text{Beh}}) \rceil.
  \]
  
\begin{figure*}[ht]
    \centering
    \begin{subfigure}{0.325\textwidth}
    	\centering
        \includegraphics[width=0.9\textwidth]{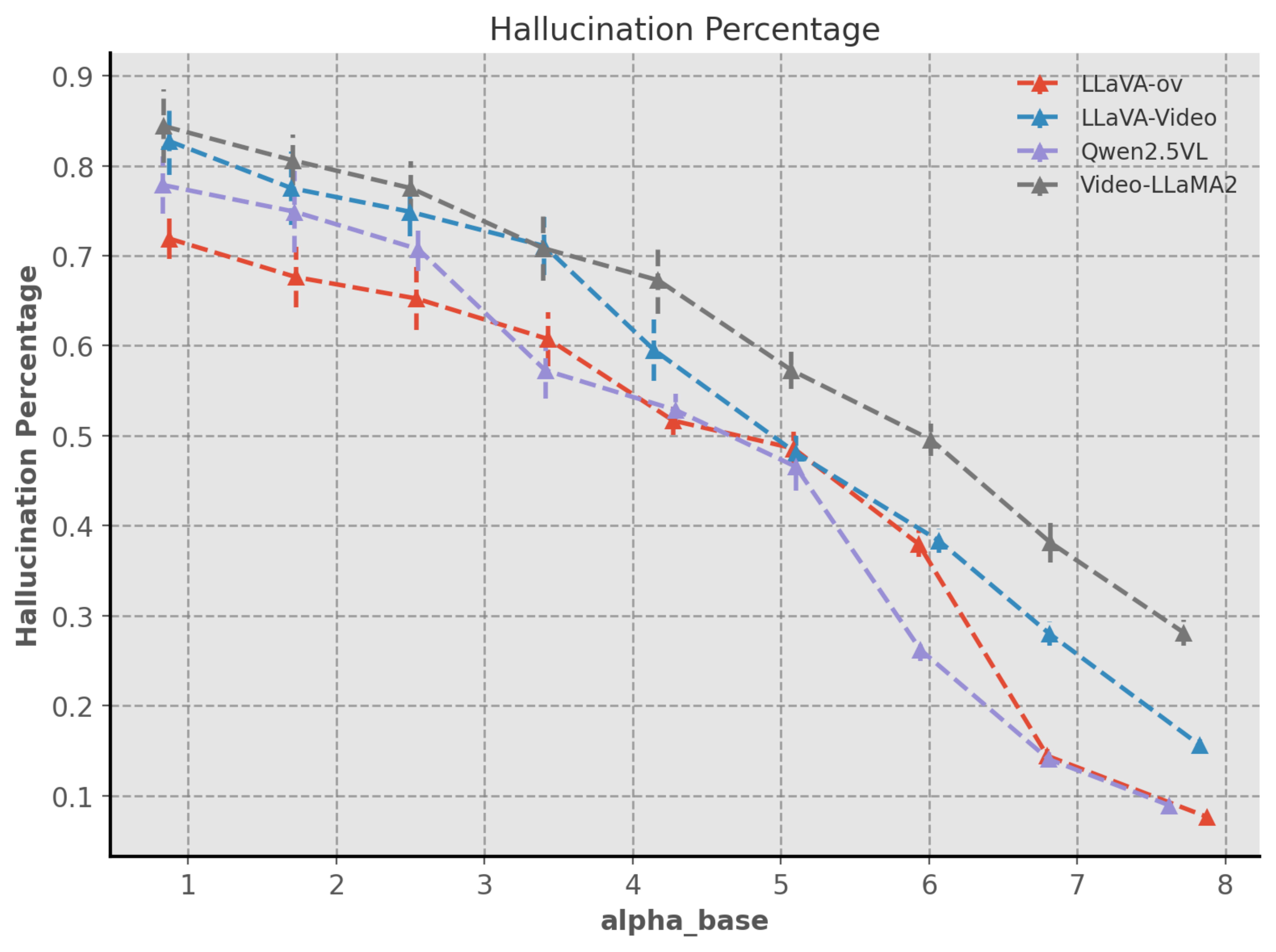}
        \caption{Hallucination Percentage}
        \label{Fig 5(a)}
    \end{subfigure}
    \hfill
    \begin{subfigure}{0.325\textwidth}
    	\centering
        \includegraphics[width=0.9\textwidth]{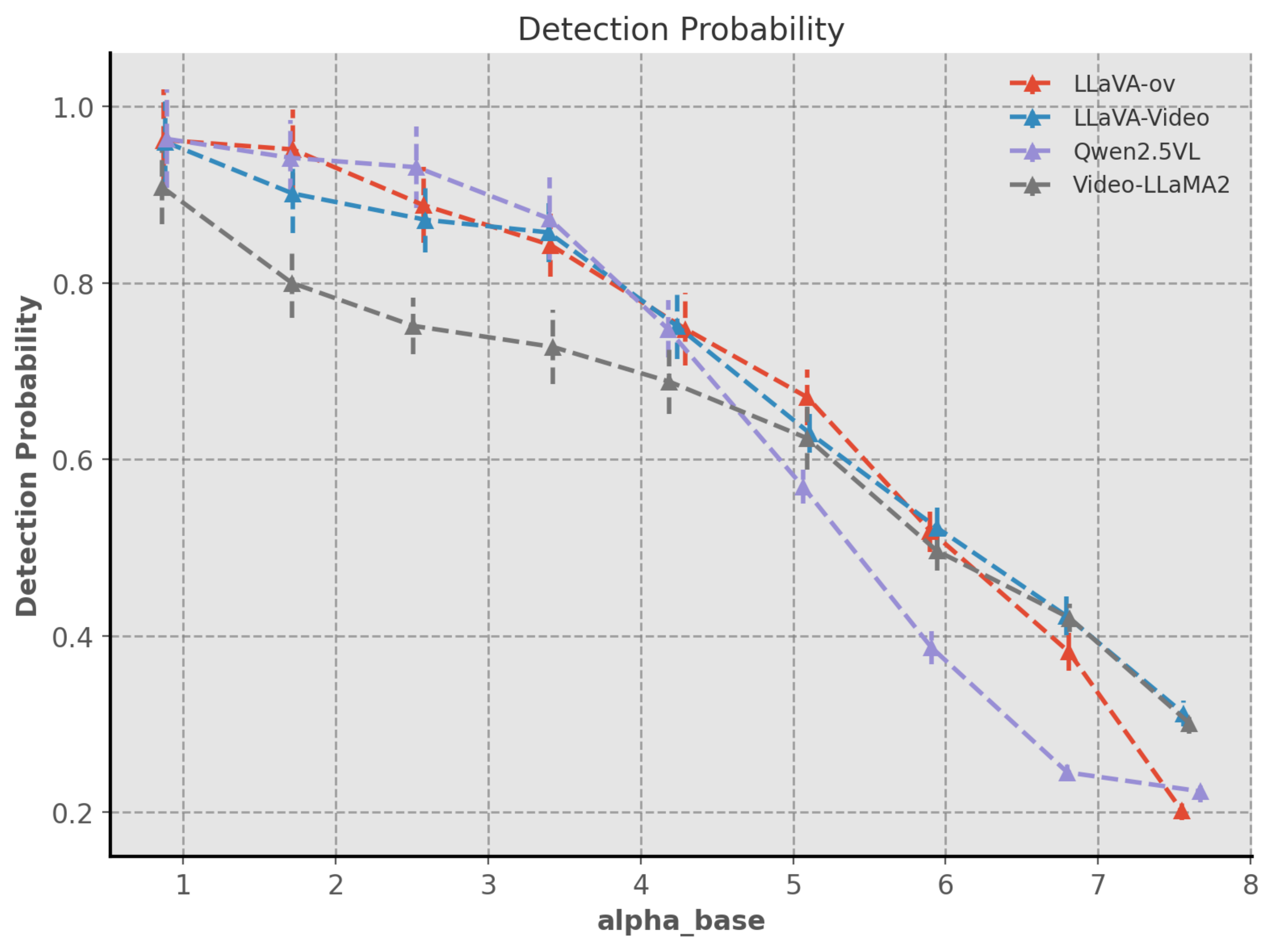}
        \caption{ Detection Probability}
        \label{Fig 5(b)}
    \end{subfigure}
    \begin{subfigure}{0.325\textwidth}
    	\centering
        \includegraphics[width=0.9\textwidth]{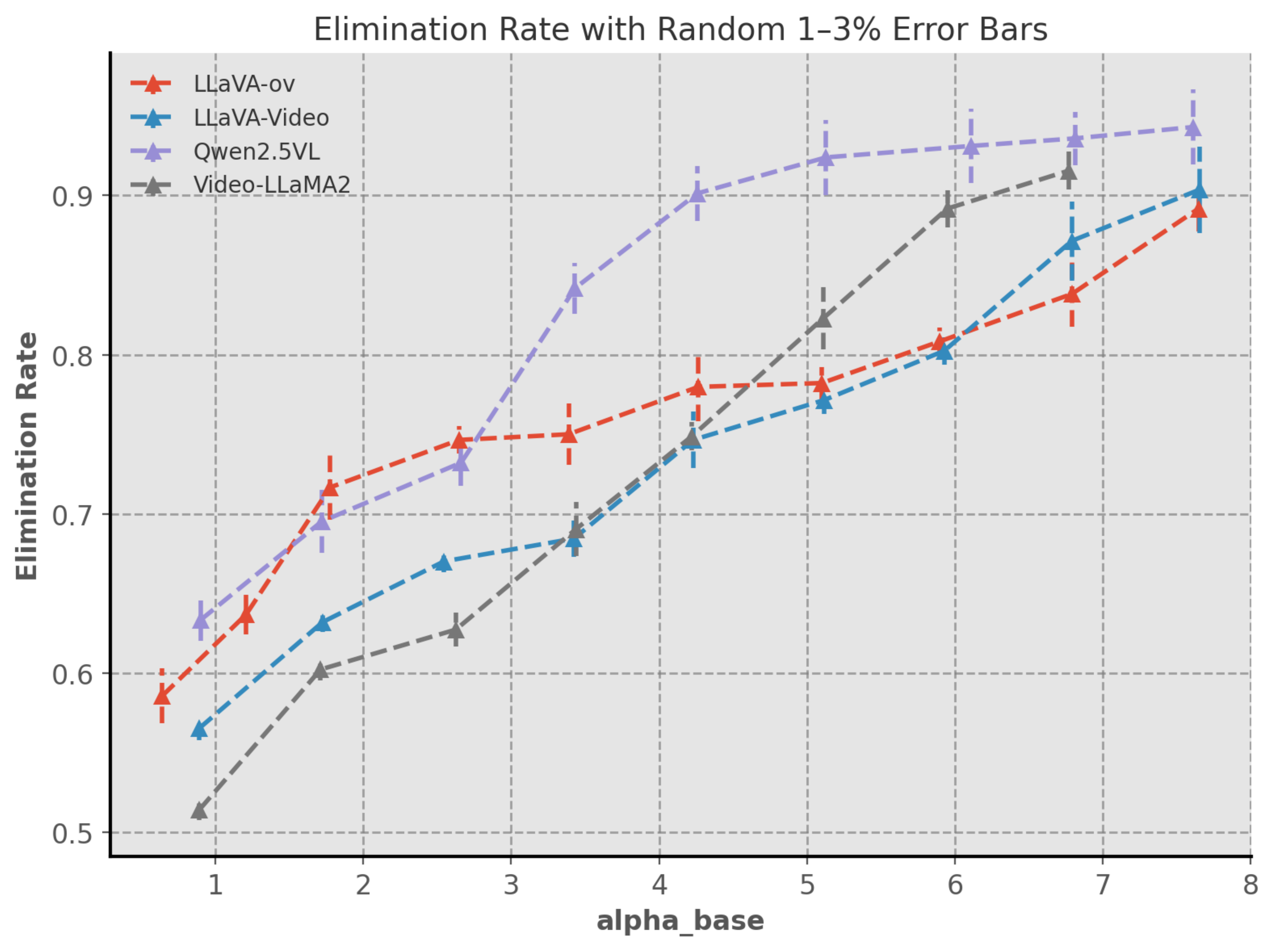}
        \caption{Elimination}
        \label{Fig 5(c)}
    \end{subfigure}
    \caption{\textbf{(a)} The probability of hallucination occurrence across the four models. \textbf{(b)} The probability of hallucinations being detected. \textbf{(c)} The probability of successful hallucination elimination.}
	\label{Fig 5}
\vspace{-12pt}
\end{figure*}

\vspace{-12pt}
\subsection{Mitigation}
The mitigation strategy directly addresses the causal chain identified in Section \ref{sec:causal_analysis} through targeted interventions. Based on the finding that behavioral hallucinations stem from prior-driven object misalignments, we first perform object grounding by orthogonalizing visual features against hallucinated object embeddings, thereby eliminating the root cause of behavioral errors before they propagate temporally.
For each hallucinated behavior \( b_i \in B_{\text{hallu}} \) identified in the previous stage, we apply the following steps:

{\bf Step 1: Extracting Behavior Text Embeddings.}
We aggregate the intermediate text-encoder embeddings of the hallucinated behavior tokens and average them to form a compact semantic vector that captures the core meaning of the erroneous behavior. We then extract the intermediate layer embedding of behavior \( b_i \) using the text encoder: $\vec{e}_i = e_{l_T}(b_i)$. 

{\bf Step 2: Correcting Image Features.}
We orthogonally project each aggregated image‐patch feature against the behavior embedding, removing the visual components that align with the hallucinated behavior. Specifically, we perform an orthogonal projection of the aggregated image feature \( e_{\text{agg}}(p_j^t) \) along the direction of the behavior text embedding at layer \( l_T \) : 
\begin{align*}
    &E_{\text{corrected}}(p_j^t) \\
    =& \left\{ E - \alpha \cdot \frac{E \cdot \vec{e}_i}{\|\vec{e}_i\|_2^2} \cdot \vec{e}_i \mid E \in E_{\text{agg}}(p_j^t) \right\}.
\end{align*}
To prevent excessive correction that could diminish genuine visual information, we limit the adjustment intensity via a correction strength coefficient \( \alpha \), which is dynamically scaled according to detection confidence:
\begin{equation*}
\alpha = \alpha_{\text{base}} \cdot (1 - \text{Confidence}_{\text{max}}(p_j^t)).   
\end{equation*} 
Intuitively, a lower confidence score, which indicates a higher risk of hallucination, should result in a stronger correction. 

{\bf Step 3: Feature Re-injection.}
We replace the original intermediate‐layer visual features with their corrected versions, ensuring that subsequent predictions rely on de‐biased, hallucination‐corrected visual representations, i.e., we replace the original intermediate layer representation \( e_{l_I}(p_j^t) \) with the corrected features \( E_{\text{corrected}}(p_j^t) \).

\begin{table*}[ht]
\centering
\caption{Behavior Hallucination Evaluation Results. 
\label{tab:behavior_hallu}}  
\resizebox{0.85\linewidth}{!}{
  \begin{tabular}{l|ccc|ccc|ccc}
    \toprule
    \multirow{2}{*}{\textbf{Method}}
      & \multicolumn{3}{c|}{\textbf{Qwen-2.5VL\textsubscript{behavior}}}
      & \multicolumn{3}{c|}{\textbf{LLaVA-ov\textsubscript{behavior}}} &  \multicolumn{3}{c}{\textbf{LLaVA-Video\textsubscript{behavior}}}\\
    \cmidrule(lr){2-4} \cmidrule(lr){5-7} \cmidrule(lr){8-10}
      & $\mathrm{BEACH\_S}\downarrow$ & $\mathrm{BEACH\_I}\downarrow$ & $\mathrm{mAP}\uparrow$
      & $\mathrm{BEACH\_S}\downarrow$ & $\mathrm{BEACH\_I}\downarrow$ & $\mathrm{mAP}\uparrow$  & $\mathrm{BEACH\_S}\downarrow$& $\mathrm{BEACH\_I}\downarrow$&$\mathrm{mAP}\uparrow$\\
    \midrule
    \rowcolor{gray!20}
    \multicolumn{10}{c}{\textbf{Momentos}} \\
    \midrule
    DECO          & 45.89 & 89.74 & 0.09 & 50.69 & 64.08 & 0.27  & 53.22& 63.37&0.28\\
    VCD           & 43.32& 59.02& 0.31& 45.50 & 61.81& 0.29  & 47.78& 64.55&0.29\\
    OPERA         & 38.88 & 57.71 & 0.30 & 53.94 & 69.13 & 0.25  & 56.68& 67.12&0.27\\
    \textbf{Ours} & 35.91& 55.33& 0.31& 42.33 & 59.49 & 0.33  & 46.92& 58.41&0.33\\
    \midrule
    \rowcolor{gray!20}
    \multicolumn{10}{c}{\textbf{SSID}} \\
    \midrule
    DECO   & 41.71 & 78.79& 0.19& 47.63& 60.57& 0.30 & 
50.01& 61.88&0.28\\
    VCD           & 34.88& 54.77& 0.35& 43.17& 59.02& 0.31 & 46.62& 55.47&
0.33\\
    OPERA         & 37.10 & 60.82& 0.29& 50.39& 65.72& 0.29 & 52.91& 67.99&
0.27\\
    \textbf{Ours} & 33.91& 53.56& 0.35 & 40.26& 57.82& 0.32 & 41.83& 53.11&
0.34\\
    \midrule
    \rowcolor{gray!20}
    \multicolumn{10}{c}{\textbf{Visual Storytelling}} \\
    \midrule
    DECO          & 36.65 & 65.81 & 0.26 & 46.47 & 68.74 & 0.26  & 
48.79& 71.68&0.23\\
    VCD           & 37.34 & 68.42 & 0.21 & 41.71 & 58.77 & 0.30  & 43.80& 59.52&
0.32\\
    OPERA         & 28.51 & 57.54 & 0.33 & 49.44 & 63.37 & 0.28  & 51.91& 57.14&
0.35\\
    \textbf{Ours} & 25.11& 53.17& 0.35& 38.80 & 54.53 & 0.32  & 40.74& 57.26&
0.35\\
    \midrule
    \rowcolor{gray!20}
    \multicolumn{10}{c}{\textbf{Visual VWP}} \\
    \midrule
    DECO          & 57.28& 74.81& 0.22& 45.77& 69.41& 0.26 & 48.43& 66.58&
0.25\\
    VCD           & 31.92& 59.76& 0.28& 41.71 & 58.77& 0.31 & 44.13& 64.79&
0.29\\
    OPERA         & 31.06& 62.94& 0.25 & 49.44 & 60.73& 0.28 & 52.30& 57.38&
0.31\\
    \textbf{Ours} & 27.43& 56.39& 0.31& 37.80& 52.59& 0.34 & 40.17& 54.11&0.35\\
    \bottomrule
  \end{tabular}
}
\vspace{-7pt}
\end{table*}
\begin{figure*}[ht]
    \centering
    \begin{subfigure}{0.325\textwidth}
    	\centering
        \includegraphics[width=0.9\textwidth]{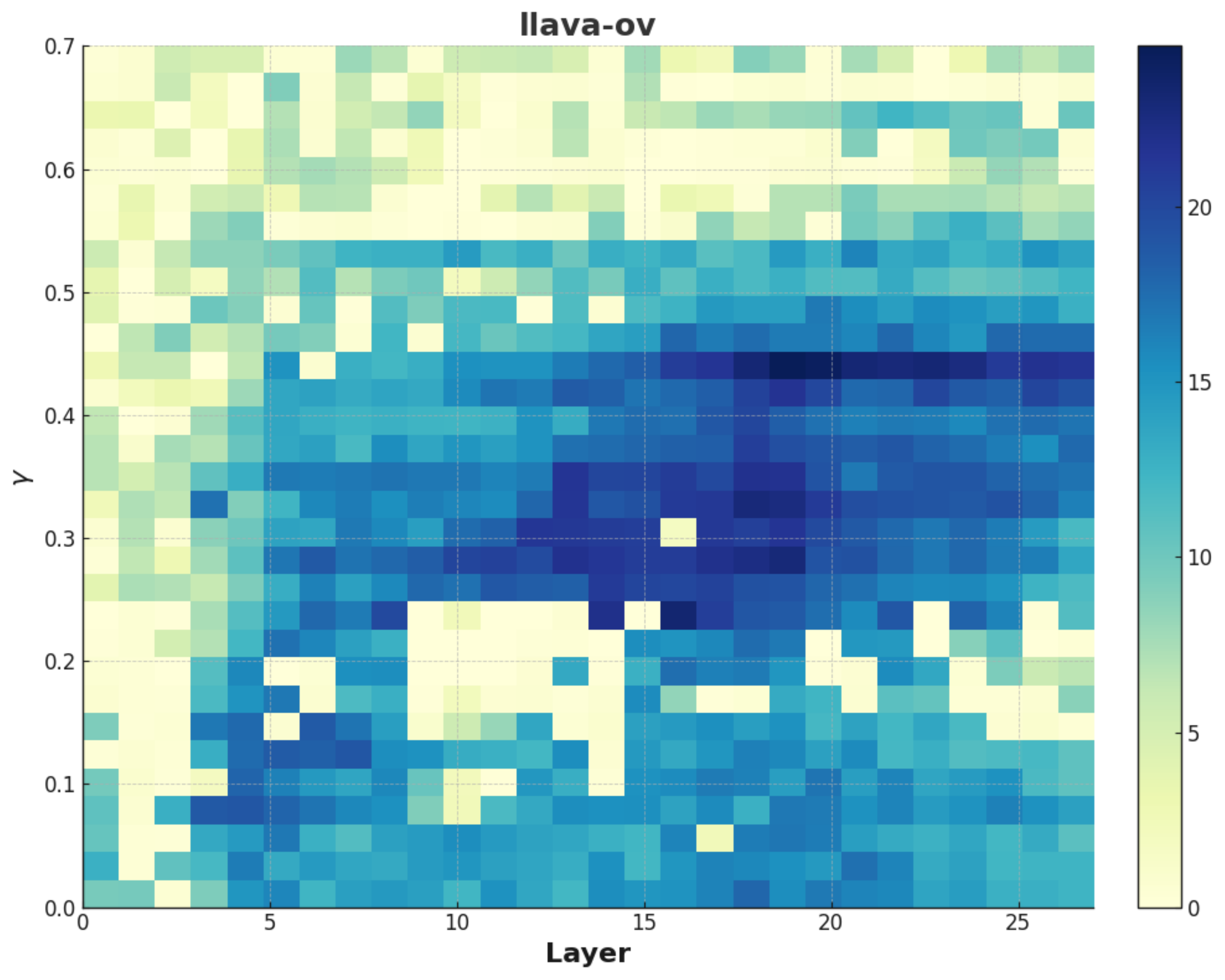}
        \caption{ Llava-OV}
        \label{Fig 6(a)}
    \end{subfigure}
    \hfill
    \begin{subfigure}{0.325\textwidth}
    	\centering
        \includegraphics[width=0.9\textwidth]{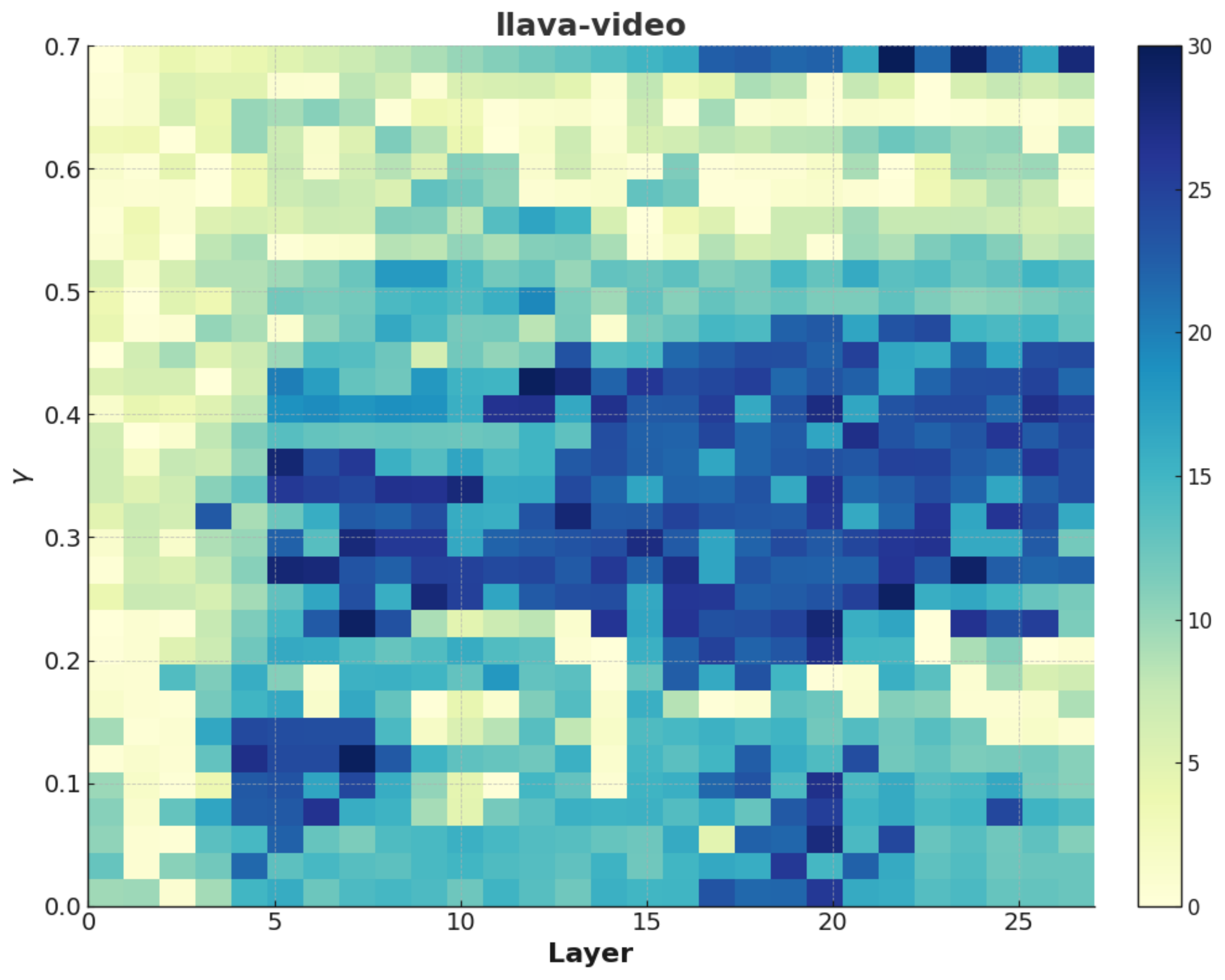}
        \caption{ Llava-video}
        \label{Fig 6(b)}
    \end{subfigure}
    \begin{subfigure}{0.325\textwidth}
    	\centering
        \includegraphics[width=0.9\textwidth]{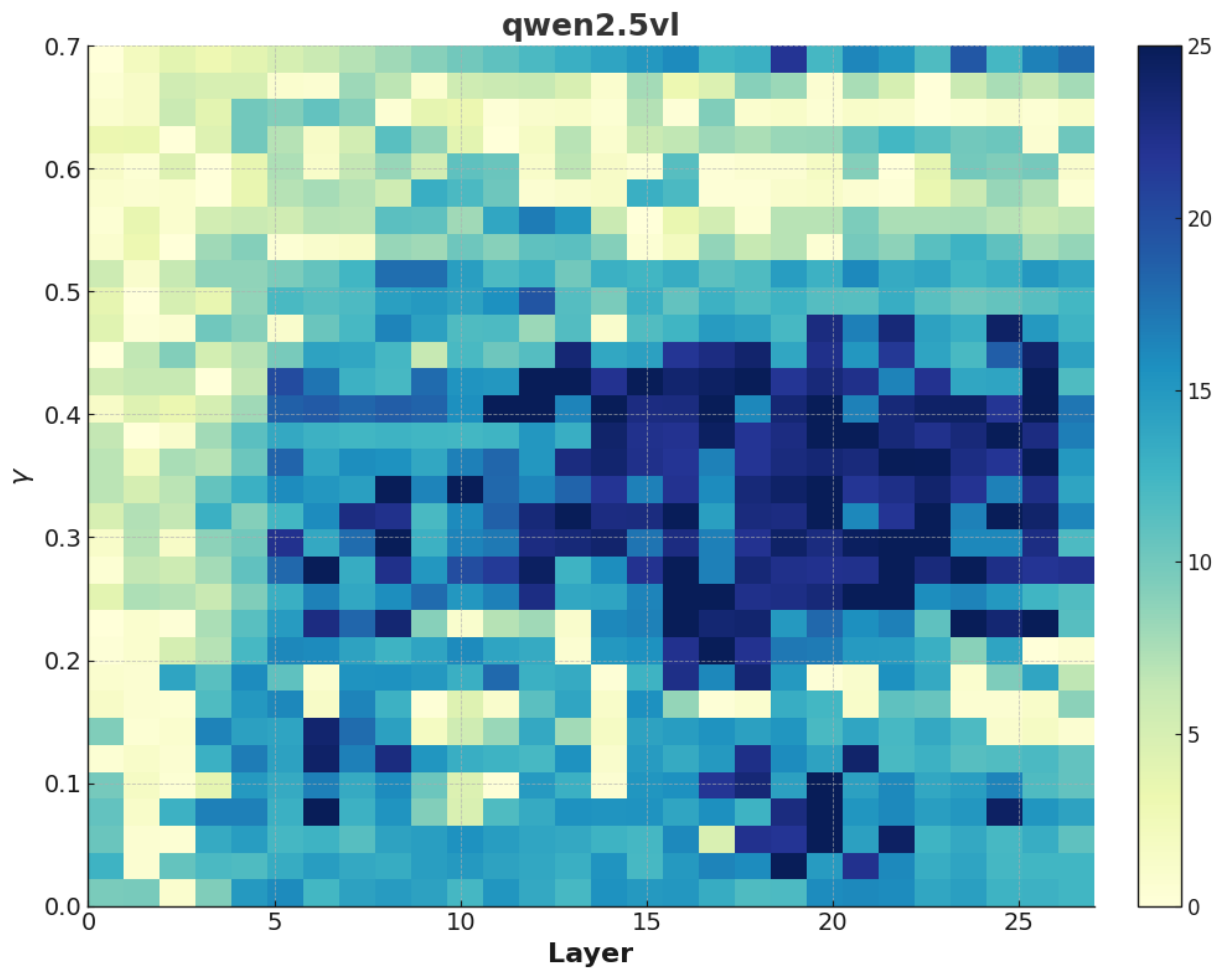}
        \caption{Qwen2.5vl}
        \label{Fig 6(c)}
    \end{subfigure}
    \caption{\textbf{(a)} Hallucination reduction across different layers and $\gamma$ value in Llava-OV model. \textbf{(b)} Hallucination reduction across different layers and $\gamma$ value in Llava-Video model. \textbf{(c)} Hallucination reduction across different layers and $\gamma$ value in Qwen-2.5vl model.}
	\label{Fig 6}
\vspace{-7pt}
\end{figure*}

\section{Experiment}
\subsection{Experimental Setup}

\noindent{\bf Datasets.} We evaluate our approach on four widely used benchmarks: \textbf{Mementos}~\citep{wang2024mementos} for sequential image captioning, \textbf{SSID}~\citep{storytelling_dataset_benchmarking} for video-based hallucination analysis, \textbf{VWP}~\citep{vwp_2023}, and \textbf{Visual Storytelling}~\citep{visual_storytelling_2016}. Details are in Appendix~\ref{Datasets}.

\noindent{\bf Baselines.} We compare our method with three state-of-the-art approaches: \textbf{DeCo}~\citep{wang2024mlm} (dynamic correction decoding), \textbf{VCD}~\citep{leng2024mitigating} (contrastive decoding), and \textbf{OPERA}~\citep{huang2023opera} (over-trust penalty). Details are in Appendix~\ref{Baselines}.

\noindent{\bf Models.} Experiments are conducted with advanced open-source multimodal large language models, including Llava-ov~\citep{li2024llavaonevision}, Qwen2.5-vl~\citep{qwen2025qwen}, Llava-Next-Video~\citep{li2024llava}, and Video-Llama~\citep{zhang2023video}.

\noindent{\bf Evaluation Metrics.} We adopt mAP and CHAIR~\citep{rohrbach2018object} as primary evaluation metrics for hallucination mitigation, in line with prior work. Additionally, we introduce \textbf{BEACH} to assess behavioral hallucinations. Details are in Appendix~\ref{finall_Metrics}.

\noindent{\bf Implementation Details.} 
In the experiments, for models based on the LLaVA architecture, we set the $\alpha_{\text{base}}$ to $[4,5]$, and $\gamma$ to $[0.4,0.6]$. For models using the Qwen-VL architecture, the beta value was set to 0.45, $\alpha_{\text{base}}$ to $[4,5]$, and $\gamma$ to $[0.4,0.6]$.


\begin{table}[htbp]
\centering
\caption{Object Hallucination on Momentos.}
\label{tab:object_hallu}
\resizebox{\linewidth}{!}{
  \begin{tabular}{@{}lcccccc@{}}
    \toprule
    \multirow{2}{*}{\textbf{Method}}
      & \multicolumn{3}{c}{\textbf{Qwen-2.5VL\textsubscript{object}}}
      & \multicolumn{3}{c}{\textbf{LLaVA-ov\textsubscript{object}}} \\
    \cmidrule(lr){2-4} \cmidrule(lr){5-7}
      & $\mathrm{CHAIR\_S}\downarrow$ & $\mathrm{CHAIR\_I}\downarrow$ & $\mathrm{mAP}\uparrow$
      & $\mathrm{CHAIR\_S}\downarrow$ & $\mathrm{CHAIR\_I}\downarrow$ & $\mathrm{mAP}\uparrow$ \\
    \midrule
    DECO          & 62.68 & 93.21 & 0.07 & 61.14 & 57.33 & 0.43 \\
    VCD           & 29.38& 38.75& 0.64& 57.79 & 55.30 & 0.44 \\
    OPERA         & 34.74& 45.35& 0.56 & 62.39 & 50.78 & 0.51 \\
    \textbf{Ours} & 31.76& 36.63& 0.65& 50.72 & 48.63 & 0.53 \\
    \bottomrule
  \end{tabular}
}
\vspace{-7pt}
\end{table}

\subsection{Experimental Results}
Results in Table~\ref{tab:behavior_hallu} demonstrate that the proposed method consistently improves hallucination suppression and behavior detection accuracy across four datasets and three model architectures. All evaluated models exhibit reductions in two BEACH metrics, accompanied by increases in mAP. Comparable but more modest improvements occur with LLaVA-ov and LLaVA-Video, reflecting their higher baseline hallucination propensity and more verbose narrative style. Qwen-2.5VL’s richer internal representations and succinct, to-the-point outputs naturally limit spurious details, making our correction mechanism especially effective; by contrast, the elaborated descriptions characteristic of the LLaVA variants introduce additional opportunities for hallucination and thus require stronger adjustments to reach similar gains.
Overall, these results confirm that our approach consistently achieves a favorable balance between hallucination mitigation and descriptive fidelity, with the degree of improvement closely tied to each model’s inherent capacity and output style. 

Table~\ref{tab:object_hallu} reports object hallucination metrics on the Momentos dataset. The performance gap between our approach and the baselines on two CHAIR metrics is minimal, demonstrating that the behavioral hallucination suppression achieved by our method does not introduce additional object hallucinations.

\subsection{Ablation Study}
\label{ablation_study}
We conducted ablation studies on the correction strength $\alpha_{\text{base}}$ and the temporal window coefficient $\gamma$, comparing fixed and dynamic strategies. Our experiments reveal that both parameters play a crucial role in balancing hallucination mitigation and output utility. Specifically, moderate values of $\alpha_{\text{base}}$ and $\gamma$—in conjunction with edits applied to middle-to-late network layers—achieve the best trade-off between hallucination reduction and descriptive fidelity. Excessively large values for either parameter can degrade output informativeness or correction accuracy. Further details and comprehensive analysis of these ablation experiments are provided in Appendix~\ref{ablation_studyinAPP}

\section{Conclusion}
This paper introduces SHE, a lightweight framework for mitigating behavioral hallucinations in multimodal large language models for sequential images. By analyzing the underlying causes, we reveal the roles of prior-driven bias and the snowball effect, and propose a two-stage solution with adaptive temporal window detection and orthogonal projection. Experiments on four benchmarks show that SHE effectively reduces behavioral hallucinations while preserving descriptive accuracy. The proposed BEACH metric also enables more precise evaluation of hallucination severity.

\section*{Limitation}
Despite the effectiveness of SHE in reducing behavioral hallucinations on a range of sequential image benchmarks, several limitations remain to be addressed. First, the framework is specifically developed for sequential images, and its applicability to other modalities such as continuous video streams or complex audio-visual data has not yet been thoroughly validated. Furthermore, the detection and mitigation steps in SHE rely on threshold-based decisions and hyperparameters, which may require careful tuning for new models or tasks to maintain optimal performance. Another limitation is the assumption of access to intermediate representations within the multimodal model. While SHE does not require additional training or model fine-tuning, it still presumes that users can extract hidden layer embeddings, which may not be feasible for proprietary, closed-source, or highly abstracted systems. However, we believe future advances in model interpretability could alleviate this constraint. We also believe SHE still offers valuable insights for controlling hallucinations in sequential visual reasoning.

\bibliography{custom}

\clearpage
\appendix

\section{More Experimental Details}
We present more experimental details and results in this section.
\subsection{Image Preprocessing}
The extracted images are preprocessed to conform to the input specifications of the Qwen2-VL model. This preprocessing stage typically involves several key steps. First, images are resized to a standardized dimension to ensure uniformity in input size, which is crucial for maintaining consistency in model performance. Second, pixel values are normalized to a specific range, usually between 0 and 1 or using z-score normalization, to optimize the model's convergence and accuracy. Additionally, other necessary transformations such as color space conversion, data augmentation techniques (e.g., rotation, flipping), and noise reduction may be applied as needed to enhance the quality and diversity of the input data.

\subsection{Caption Generation}
The preprocessed images are fed into the Qwen2-VL model to generate textual captions. During this process, each image is individually processed by the model, which leverages its vision and language components to produce a coherent and contextually appropriate caption. The model's output is recorded for subsequent analysis, ensuring that the generated captions are both accurate and representative of the visual content.

\subsection{Co-occurrence Score Calculation} 
For each generated caption, a Co-occurrence Score is computed to evaluate the co-occurrence patterns of objects and behaviors within the text. This score quantifies how frequently and in what manner different objects and behaviors appear together in the captions. Unlike traditional object-object co-occurrence metrics, the calculation here is adapted to focus on the relationships between objects and behaviors, as well as among behaviors themselves. This adjustment allows for a more nuanced understanding of how well the model captures the interactions and contextual relationships within the images, providing valuable insights into the model's performance in representing real-world scenarios accurately. 
\begin{table}[ht]
    \centering
    \caption{Model Performance Comparison.}
    \resizebox{\linewidth}{!}{
        \begin{tabular}{l|c|c|c|c}
            \toprule
            \textbf{Model} & \textbf{Recall} & \textbf{Precision} & \textbf{F1} & \textbf{Correlation} \\ \midrule
            GPT-4V & 0.120 & 0.180 & 0.132  & 0.23 \\ 
            Gemini & 0.165 & 0.179 & 0.146  & 0.35 \\ 
            Video-LLaMA-2 & 0.197 & 0.067 & 0.125  & 0.19 \\ 
            Chat-UniVi & 0.127 & 0.184 & 0.172  & 0.27 \\ 
            LLaVA-Video & 0.112 & 0.134 & 0.106  & 0.41 \\ 
            MiniGPT4 & 0.135 & 0.145 & 0.115  & 0.45 \\ 
            mPLUG-Owl-v3 & 0.106 & 0.113 & 0.069  & 0.17 \\ 
            InstructBLIP & 0.133 & 0.125 & 0.127  & 0.36 \\ \bottomrule
        \end{tabular}
    }
\vspace{-7pt}
\end{table}
\begin{figure}[ht]
    \centering
   \includegraphics[width=\linewidth]{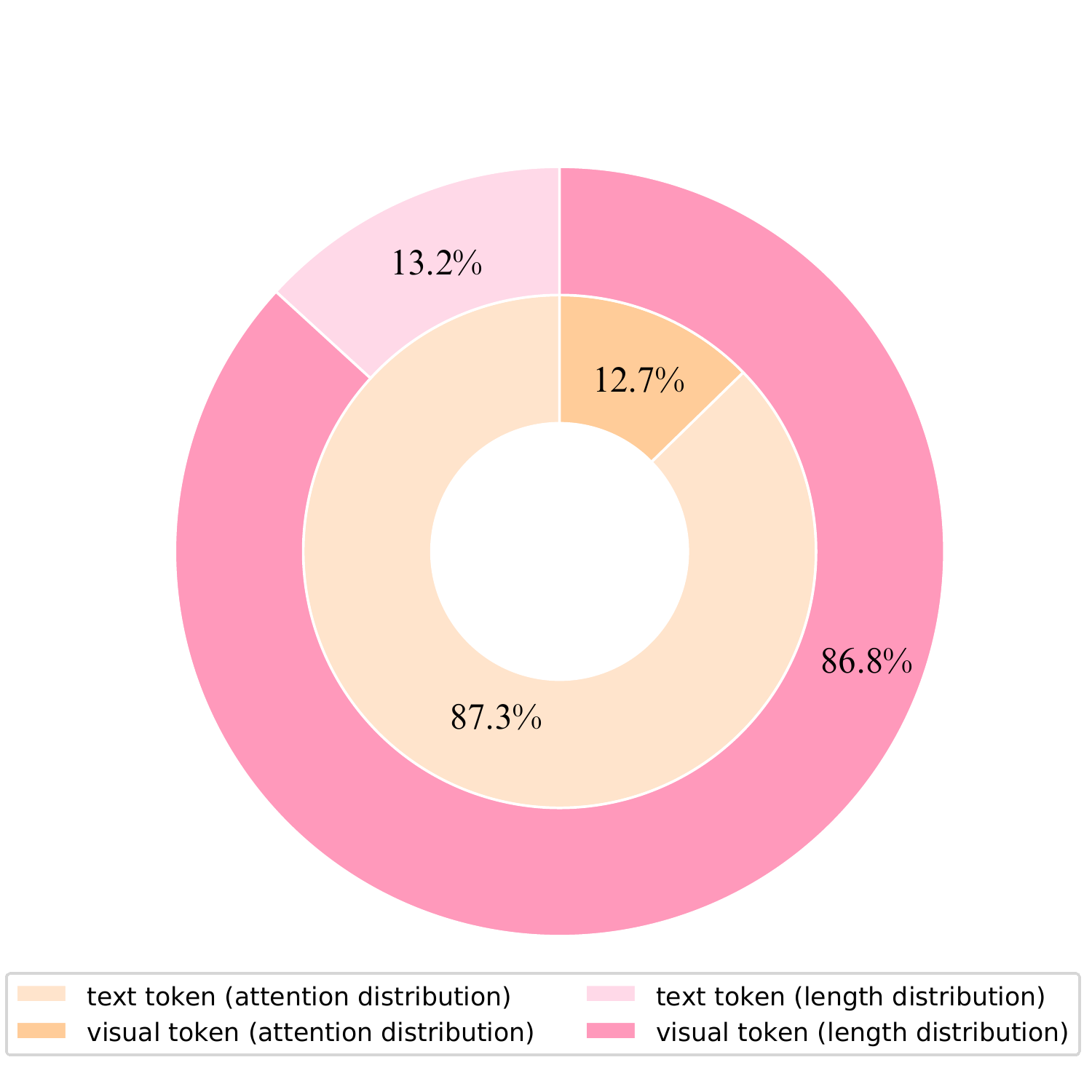}
    \caption{Attention distribution.}
    \label{Fig 3}
\end{figure}
\begin{table*}[htbp]
\centering
\caption{Behavior Hallucination Evaluation Results. \label{tab:behavior_hallu2} }
\resizebox{0.87\linewidth}{!}{
  \begin{tabular}{l|ccc|ccc|ccc}
    \toprule
    \multirow{2}{*}{\textbf{Method}}
      & \multicolumn{3}{c|}{\textbf{Qwen-2.5VL\textsubscript{behavior}}}
      & \multicolumn{3}{c|}{\textbf{LLaVA-ov\textsubscript{behavior}}}
      & \multicolumn{3}{c}{\textbf{LLaVA-Video\textsubscript{behavior}}} \\
    \cmidrule(lr){2-4} \cmidrule(lr){5-7} \cmidrule(lr){8-10}
      & $\mathrm{BEACH\_S}\downarrow$ & $\mathrm{BEACH\_I}\downarrow$ & $\mathrm{mAP}\uparrow$
      & $\mathrm{BEACH\_S}\downarrow$ & $\mathrm{BEACH\_I}\downarrow$ & $\mathrm{mAP}\uparrow$
      & $\mathrm{BEACH\_S}\downarrow$ & $\mathrm{BEACH\_I}\downarrow$ & $\mathrm{mAP}\uparrow$ \\
    \midrule
    \rowcolor{gray!20}
    \multicolumn{10}{c}{\textbf{Momentos}} \\
    \midrule
    DECO          & 49.56 & 91.53 & 0.09 & 54.75 & 69.21 & 0.26 & 57.48 & 68.44 & 0.25 \\
    VCD           & 46.79 & 63.74 & 0.29 & 49.14 & 66.75 & 0.28 & 51.60 & 69.71 & 0.29 \\
    OPERA         & 41.99 & 62.33 & 0.28 & 58.26 & 74.66 & 0.24 & 61.21 & 72.49 & 0.27 \\
    \textbf{Ours} & 38.78 & 59.76 & 0.29 & 45.72 & 64.25 & 0.31 & 50.67 & 63.08 & 0.31 \\
    \midrule
    \rowcolor{gray!20}
    \multicolumn{10}{c}{\textbf{SSID}} \\
    \midrule
    DECO          & 45.05 & 82.73 & 0.18 & 51.44 & 65.42 & 0.28 & 54.01 & 66.83 & 0.26 \\
    VCD           & 37.67 & 59.15 & 0.33 & 46.62 & 63.74 & 0.29 & 50.35 & 59.91 & 0.30 \\
    OPERA         & 40.07 & 65.69 & 0.28 & 54.42 & 70.98 & 0.28 & 57.14 & 73.43 & 0.24 \\
    \textbf{Ours} & 36.62 & 57.84 & 0.33 & 43.48 & 62.45 & 0.30 & 45.18 & 57.36 & 0.33 \\
    \midrule
    \rowcolor{gray!20}
    \multicolumn{10}{c}{\textbf{Visual Storytelling}} \\
    \midrule
    DECO          & 39.58 & 71.07 & 0.25 & 50.19 & 74.24 & 0.25 & 52.69 & 75.26 & 0.22 \\
    VCD           & 40.33 & 73.89 & 0.20 & 45.05 & 63.47 & 0.28 & 47.30 & 64.28 & 0.30 \\
    OPERA         & 30.79 & 62.14 & 0.31 & 53.40 & 68.44 & 0.27 & 56.06 & 61.71 & 0.31 \\
    \textbf{Ours} & 27.12 & 57.42 & 0.33 & 41.90 & 58.89 & 0.30 & 44.00 & 58.84 & 0.33 \\
    \midrule
    \rowcolor{gray!20}
    \multicolumn{10}{c}{\textbf{Visual VWP}} \\
    \midrule
    DECO          & 61.86 & 78.55 & 0.21 & 49.43 & 74.96 & 0.25 & 52.30 & 71.91 & 0.25 \\
    VCD           & 34.47 & 64.54 & 0.27 & 45.05 & 63.47 & 0.29 & 47.66 & 69.97 & 0.27 \\
    OPERA         & 33.54 & 67.98 & 0.24 & 53.40 & 65.59 & 0.27 & 56.48 & 61.97 & 0.28 \\
    \textbf{Ours} & 29.62 & 60.90 & 0.29 & 40.82 & 56.80 & 0.32 & 43.38 & 58.44 & 0.32 \\
    \bottomrule
  \end{tabular}
}
\end{table*}

\paragraph{Video-MME.}
We ensure that each group contains a diverse set of videos from different domains and sub-domains to maintain the representativeness of the dataset. \\
For each video in the dataset, we utilize the provided question-answer pairs to prompt the models and generate responses. The process begins with question generation, where a set of questions covering various aspects of video understanding, such as object recognition, behavior analysis, and scene comprehension, is created based on the Video-MME dataset's question-answer pairs. Next, in the model inference stage, each video is input into the selected MLLMs, which are then asked to answer these generated questions, producing natural language responses. The generated responses are subsequently compared against the ground-truth answers from the dataset. Here, we employ a combination of exact match and semantic similarity metrics to assess the correctness of the model's response. If the model's response contains objects or behaviors not present in the video or the standard answers, it is flagged as a hallucination, which we further categorize into object hallucinations and behavior hallucinations.

\begin{table*}[ht]
\centering
\caption{VideoMME Example Data.}
\label{tab:example_data}
\resizebox{0.7\linewidth}{!}{
\begin{tabular}{|c|c|c|p{3cm}|p{3cm}|p{2.5cm}|c|}
\toprule
\textbf{Video ID} & \textbf{Duration} & \textbf{Type} & \textbf{Subtitle Content} & \textbf{Question} & \textbf{Options} & \textbf{Answer} \\ 
\midrule
001 & 1 min & Edu & Hi guys, I'm going to show you how to prepare a ... & Person's clothing color? & A. Black B. Gray C. Green D. Brown & C \\ 
\midrule
002 & 2 min & Doc & Exploring ancient ruins in the jungle... & Main subject? & A. Modern city B. Ancient ruins C. Jungle animals D. Local cuisine & B \\ 
\midrule
003 & 3 min & Tut & Starting by gathering materials... & First step? & A. Clean workspace B. Gather materials C. Turn on machine D. Put on gloves & B \\ 
\bottomrule
\end{tabular}}
\end{table*}

\subsection{Datasets}
\label{Datasets}

\paragraph{Mementos.} The Mementos dataset is a comprehensive benchmark designed to evaluate the sequential image reasoning abilities of Multimodal Large Language Models (MLLMs). We sample 5000 sequence images from the Mementos~\citep{wang2024mementos}, using the image captioning objective to caption methods with both Qwen2.5-vl and LLaVA-ov. 

\paragraph{SSID.} SSID~\citep{storytelling_dataset_benchmarking} comprises 17,365 video frames organized into 3,473 chronological 5-image sequences, each paired with four human-written, 5-sentence stories (13,892 total) to benchmark expressive and coherent visual storytelling models.

\paragraph{VWP.} VWP~\citep{vwp_2023} comprises nearly 2,000 curated sequences of movie shots, each containing 5–10 images selected to form coherent, character‐focused visual narratives. 
 
\paragraph{Visual Storytelling.} Visual Storytelling~\citep{visual_storytelling_2016} comprises 210,819 unique Flickr photos organized into 50,000 five-image sequences (albums of 10–50 images captured within a 48-hour span), each annotated by Amazon Mechanical Turk workers with a coherent five-sentence story (one sentence per image) to benchmark sequential vision-to-language narrative generation models.

 

\subsection{Metrics}
\label{finall_Metrics}
The metrics we used, such as Recall, Precision, and F1 Score, were used to measure the models' understanding of image sequences. Additionally, for hallucination detection, we use the Mean Average Precision (mAP) metric and CHAIR metric~\citet{rohrbach2018object}. 
In addition to this, we are the first to propose Behavior Evaluation Assessment for Caption Hallucination (BEACH), a metric used to assess behavioral hallucinations.
\begin{equation*}
\text{BEACH}_I = \frac{|\{\text{hallucinated behaviors}\}|}{|\{\text{all annotated behaviors}\}|}, 
\end{equation*}
\begin{equation*}
\text{BEACH}_S 
= \frac{|\{\text{captions with hallucinated behaviors}\}|}{|\{\text{all captions}\}|}.     
\end{equation*}

\subsection{Baselines} 
\label{Baselines}
\paragraph{Deco.} DeCo~\citep{wang2024mlm}, a dynamic correction decoding method for reducing hallucinations in MLLMs. DeCo adaptively selects preceding layers and integrates knowledge into the final layer to adjust output logits, effectively mitigating hallucinations.

\paragraph{VCD.} VCD~\citep{Leng2023Mitigating} is a method designed to mitigate object hallucinations in large vision-language models (LVLMs) by leveraging contrastive decoding.
 
\paragraph{OPERA.} OPERA~\citep{huang2023opera} is a novel decoding method that introduces an Over-trust Penalty and a Retrospection-Allocation strategy to mitigate hallucinations without additional data, knowledge, or training. 

\subsection{Ablation Study}
\label{ablation_studyinAPP}

\paragraph{Ablation on Correction Strength $\alpha_{\text{base}}$.}
We compare fixed $\alpha_{\text{base}}$ [1,6] versus dynamic  $\alpha_{\text{base}}$ strategies (same value range) using Mementos's test set containing 100 manually annotated over-correction and under-correction cases. 
We vary the $\alpha_{\text{base}}$ and measure changes in the rate of detected hallucinatory behavior, rate of hallucinatory behavior, and rate of elimination of hallucinatory behavior. Figure~\ref{Fig 5} shows that as the $\alpha_{\text{base}}$ increases, our method is getting better at removing the hallucination, but by looking at the actual model output and analyzing it, we found that the model output becomes uninformative and useless, when the $\alpha_{\text{base}}$ is too high. 

\paragraph{Ablation on Temporal Window Coefficient  $\gamma$.}
We investigate how the window coefficient $\gamma$ [0.2,0.8] affects behavior complexity adaptation. For each value of $\gamma$, we analyze correction success rates, and we monitor the reduction in hallucination across various layers designated for editing. We extract latent embeddings for the text embedding and exclude the parameters that lead to a decrease in correctly identified behavior. Pearson correlation quantifies the coefficient-accuracy relationship. When $\gamma$ takes on an appropriate value—typically in the range of 0.4 to 0.6—and the targeted layer lies in the middle-to-late stages of the network, our method achieves a favorable balance between eliminating hallucinations and preserving the validity of the model’s outputs. If $\gamma$ becomes too large, the removal accuracy (as shown in the figure) deteriorates; likewise, a layer-wise analysis reveals that applying the removal at either the earliest or the very last layers yields inferior results compared to intervening in the mid-to-late layers.

\end{document}